\def\paperlanguage{}
\pdfoutput=1 

\documentclass[letterpaper, 10 pt, conference]{ieeeconf}  

\usepackage{bm}
\usepackage{cite}
\usepackage{flushend}
\include{preamble}

\IEEEoverridecommandlockouts                              

\title{\LARGE \textbf
  {
    \switchlanguage%
    {%
      Long-time Self-body Image Acquisition and its Application to the Control of Musculoskeletal Structures
    }%
    {%
      筋骨格構造における長期的自己身体像獲得と制御機構への応用
    }%
  }
}

\author{Kento Kawaharazuka$^{1}$, Kei Tsuzuki$^{1}$, Shogo Makino$^{1}$, Moritaka Onitsuka$^{1}$\\Yuki Asano$^{1}$, Kei Okada$^{1}$, Koji Kawasaki$^{2}$, and Masayuki Inaba$^{1}$
  \thanks{$^{1}$ The authors are with Department of Mechano-Informatics, Graduate School of Information Science and Technology, The University of Tokyo, 7-3-1 Hongo, Bunkyo-ku, Tokyo, 113-8656, Japan.
    {\texttt\small [kawaharazuka, tsuzuki, makino, onitsuka, asano, k-okada, inaba]@jsk.t.u-tokyo.ac.jp}
    }
  \thanks{$^{2}$ The author is associated with TOYOTA MOTOR CORPORATION.
    {\texttt\small koji\_kawasaki@mail.toyota.co.jp}
  }
}
\begin{document}

\maketitle
\thispagestyle{empty}
\pagestyle{empty}

\begin{abstract}
  \switchlanguage%
  {%
    The tendon-driven musculoskeletal humanoid has many benefits that human beings have, but the modeling of its complex muscle and bone structures is difficult and conventional model-based controls cannot realize intended movements.
    Therefore, a learning control mechanism that acquires nonlinear relationships between joint angles, muscle tensions, and muscle lengths from the actual robot is necessary.
    In this study, we propose a system which runs the learning control mechanism for a long time to keep the self-body image of the musculoskeletal humanoid correct at all times.
    Also, we show that the musculoskeletal humanoid can conduct position control, torque control, and variable stiffness control using this self-body image.
    We conduct a long-time self-body image acquisition experiment lasting 3 hours, evaluate variable stiffness control using the self-body image, etc., and discuss the superiority and practicality of the self-body image acquisition of musculoskeletal structures, comprehensively.
  }%
  {%
    筋骨格ヒューマノイドは人体模倣型ゆえの利点を多く有すると同時に, 複雑な筋構造・骨格構造を持つためモデル化が難しく, 従来のモデルベース制御では意図した動作は難しい.
    そのため, 実機動作から関節-筋張力-筋長の非線形な関係を獲得する学習機構が必要となり, 本研究ではこの学習機構を長期的に走らせ, 常に自己身体像を正しく保つための仕組みを提案する.
    また, この関節-筋空間マッピングを利用することで, 関節の位置制御・力制御を行うことができると同時に, 筋骨格構造の特徴である, 非線形弾性要素と拮抗構造を利用した可変剛性制御が可能であることを示す.
    本研究では, 3時間に及ぶ自己身体像の獲得実験・自己身体像を用いた可変剛性制御の実機検証等を行い, 筋骨格構造における自己身体像獲得の優位性と実用性について総括的に議論する.
  }%
\end{abstract}

\section{INTRODUCTION}\label{sec:introduction}
\switchlanguage%
{%
  The tendon-driven musculoskeletal humanoid \cite{nakanishi2013design, wittmeier2013toward, asano2016kengoro} has many benefits that human beings have, such as multiple degrees of freedom (multi-DOFs), under-actuated structures of the spine and fingers, variable stiffness control using nonlinear elasticity and antagonism of muscles, and error correction using redundant muscles.
  Therefore, the humanoid is expected to move flexibly like human beings and work in an environment with physical contact.
  At the same time, its bone and muscle structures are complex, and there are many problems which cannot be solved by conventional model-based controls.
  In particular, unlike the tendon-driven robot with constant moment arm \cite{hirose1991tendon, marques2013myorobotics}, the muscle route modeling of the musculoskeletal humanoid is very difficult.

  In order to solve these problems, various studies have been conducted.
  There is the method which trains the neural network of the joint-muscle mapping (JMM, the nonlinear relationship between joint angles and muscle lengths) with the actual sensor information \cite{mizuuchi2006acquisition}, the method which trains JMM using polynomial regression and estimates the current joint angles from JMM \cite{ookubo2015learning}, and the method which obtains the Jacobian between the position and muscle length from vision \cite{motegi2012jacobian}.
  Also, we have developed an online learning method of JMM using vision \cite{kawaharazuka2018online}.
  By extending it, we have also developed an online learning method of the self-body image considering muscle-route changes caused by body tissue softness \cite{kawaharazuka2018bodyimage}.

  However, because there is a large error in the joint-muscle mapping between the actual robot and its geometric model in the early stages of learning, muscle temperatures rise rapidly due to unintended high muscle tensions, and the motors of the muscles may burn out.
  Also, previous studies conducted the online learning for only five minutes, and long-time online learning sometimes proceeds in an unintended direction.
  So, we propose a mechanism to stably conduct long-time self-body image acquisition.
  This mechanism includes a simple online updater of the self-body image realized by separating software elasticity and hardware elasticity, data accumulation and augmentation of the actual robot sensor information for online learning, and a safety mechanism considering muscle tension and muscle temperature.

  Also, previous studies have focused on the learning of the self-body image itself, so there are few studies on control methods using it.
  In this study, we develop the position and torque control using the acquired self-body image.
  Additionally, the musculoskeletal humanoid can conduct mechanical variable stiffness control, using its redundant muscles and the nonlinear elastic element of each muscle.
  Until now, although model-based variable stiffness control systems such as \cite{kobayashi1998tendon} have been developed, these control systems can be used only when the moment arm and nonlinear elasticity of muscles are modelized completely.
  Therefore, we propose an estimation of mechanical operational stiffness using the self-body image, and its control which enables the change of the operational stiffness as intended.
  These proposed methods will extend the range of application of musculoskeletal humanoids.

  This study is important for not only the tendon-driven musculoskeletal humanoid, but also for the musculoskeletal hand, such as the ACT Hand \cite{weghe2004acthand}, the tensegrity robot \cite{paul2006tensegrity}, soft robotics \cite{niiyama2010athlete}, etc.

  In the following sections, first, we will state the overview of the musculoskeletal structure.
  Then, we will explain the detailed method of long-time self-body image acquisition and control systems using it, while comparing with previous studies.
  Finally, we will conduct several experiments of the proposed methods, and state the conclusion.
}%
{%
  筋骨格ヒューマノイド\cite{nakanishi2013design, wittmeier2013toward, jantsch2013anthrob, asano2016kengoro}は人間のような多自由度, 背骨や指の劣駆動機構, 筋の非線形弾性と拮抗構造を利用した可変剛性制御, 冗長な筋配置によるエラー訂正等の利点を有しており, 人体を模倣しているがゆえに人間と同様な柔らかい動作を行うことができると期待されている.
  同時に, その骨格や筋構成は複雑であり, 現在主流である幾何モデルをベースとした制御だけでは解決できない問題が多数存在する.
  特に, モーメントアームが一定でモデル化が容易な腱駆動型ロボット\cite{hirose1991tendon, marques2013myorobotics}とは違い, 筋骨格ヒューマノイドの筋経路は単純なモデル化が難しいことが大きな要因である.

  そららの問題を解決するために, これまで様々な研究が成されてきた.
  動きのセンサ情報からニューラルネットワークで表した関節-筋空間マッピングを学習させる方法\cite{mizuuchi2006acquisition}, 多項式近似を用いて関節-筋空間マッピングを表しそれを微分して用いることで関節角度推定を行う方法\cite{ookubo2015learning}, 視覚から位置と筋長の間のヤコビアンを取得していく方法\cite{motegi2012jacobian}等が存在する.
  また, ニューラルネットワークで表現した関節-筋空間マッピングを, 視覚を用いてオンラインに学習する方法が開発されている\cite{kawaharazuka2018online}.
  さらにそれを発展させ, 関節-筋張力-筋長の関係を学習することで, 筋骨格ヒューマノイド特有の身体組織の柔軟性を考慮した自己身体像のオンライン学習が可能となってきた\cite{kawaharazuka2018bodyimage}.

  しかし, 学習初期において幾何モデルと実機の関節-筋空間マッピングは大きく異なり, 意図しない高張力によって激しく筋モータの温度が上昇し, モータが焼損してしまうことが多々ある.
  また, \cite{kawaharazuka2018bodyimage}では5分程度しかオンライン学習を実行していなく, 長時間学習を稼働させてしまうと学習が意図しない方向に過学習してしまうこと等が散見された.
  そこで我々は, 長期的に自己身体像を獲得するための, 自己身体像の構築手法, オンライン学習の工夫や安全動作機構を提案する.
  これには, ソフトウェアばねとハードウェアばねを分離することによる簡潔な自己身体像の逐次更新・オンライン学習における実機データの蓄積と拡張・筋張力と筋温度を考慮した安全機構等の工夫が含まれる.

  また, これまで自己身体像を学習することに論点が置かれ, それらを用いた制御手法の開発は乏しかった.
  本研究では, 獲得された自己身体像を用いた位置制御・トルク制御手法を開発する.
  また, 筋骨格ヒューマノイドはその冗長な筋群と各筋の非線形弾性を利用することで, ハードウェアによる可変剛性制御を実現することができるとされる.
  今まで, \cite{kobayashi1998tendon}のような研究は多くなされてきたが, モーメントアームや非線形弾性要素等が完全にモデル化できる場合の制御であり, 本研究では入力に対する出力のみが得られるニューラルネットワークによって表現された身体モデルを使って, どのようにハードウェア剛性を導出・変化させることができるかを示す.
  そこで, 得られた自己身体像を用いて筋骨格ヒューマノイドにおける作業空間ハードウェア剛性の推定と, 作業空間剛性の任意操作を可能とする制御を提案する.
  これらにより, 筋骨格ヒューマノイドの行動範囲がより広がることと確信する.

  本研究は筋骨格ヒューマノイドだけではなく, ACT HAND \cite{weghe2004acthand}のような筋骨格構造を持つハンドや, Tensegrity Robot \cite{paul2006tensegrity}, 空気圧アクチュエータにより動作するソフトロボット\cite{niiyama2010athlete}等の研究に対しても大きな意味があると考える.

  後の章ではまず, 筋骨格ヒューマノイドの身体構造の概要について説明する.
  次に, 先行研究と比較しながら, 長期的自己身体像獲得手法について述べ, 獲得された自己身体像を用いた制御手法について説明する.
  最後に, 提案手法に関する実験を行い, 結論を述べる.
}%

\begin{figure*}[t]
  \centering
  \includegraphics[width=1.8\columnwidth]{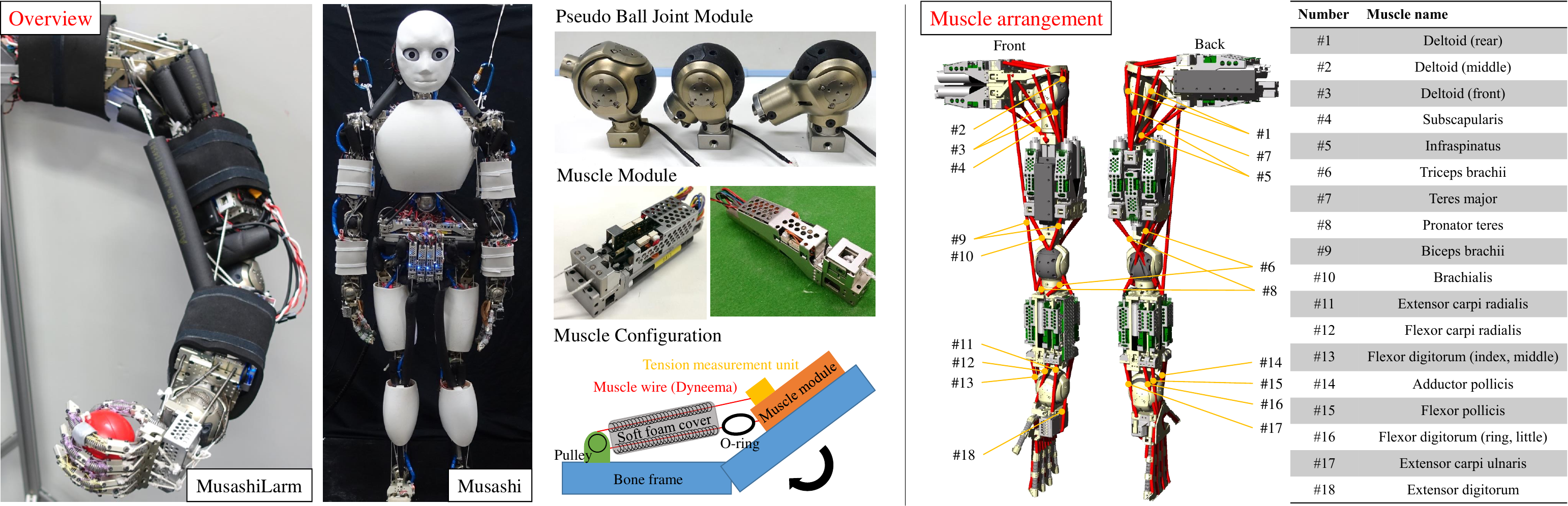}
  \caption{Overview of newly developed MusashiLarm and Musashi used in this study \cite{kawaharazuka2018musashilarm-en}.}
  \label{figure:musculoskeletal-structure}
  \vspace{-3.0ex}
\end{figure*}

\section{Musculoskeletal Humanoid} \label{sec:musculoskeletal-structure}
\switchlanguage%
{%
  In this study, we use MusashiLarm and Musashi \cite{kawaharazuka2018musashilarm-en} (\figref{figure:musculoskeletal-structure}) developed as a musculoskeletal research platform to succeed Kengoro \cite{asano2016kengoro}.
}%
{%
  本研究では腱悟郎\cite{asano2016kengoro}の研究開発用後継機として作られたMusashiLarmとMusashi \cite{kawaharazuka2018musashilarm-en} (\figref{figure:musculoskeletal-structure})を用いて実験を行う.
}%

\subsection{Joint Structure} \label{subsec:joint-configuration}
\switchlanguage%
{%
  MusashiLarm has 3 DOFs of the shoulder, 2 DOFs of the elbow and radioulnar joint, 2 DOFs of the wrist joint, and the fingers have flexible and robust under-actuated structures made of machined springs.
  Each joint is constructed by a pseudo ball joint module which can measure joint angles directly using the included potentiometers.
  While ordinary musculoskeletal humanoids cannot include joint angle sensors due to ball joints, by this configuration, we made the experimental evaluation easy.
  Among these joints, we mainly consider 3 DOFs of the shoulder and 2 DOFs of the elbow in this study.
  Musashi is a simple extension of MusashiLarm, and we use the dual arms of Musashi in several experiments.
}%
{%
  本研究で用いる筋骨格上肢は肩関節の3自由度, 肘関節と橈尺関節の2自由度, 手首関節の2自由度を持ち, 指は劣駆動型の切削ばねによる柔軟かつ頑健な構造\cite{makino2018hand}を持つ.
  これら関節の3自由度, 2自由度, 2自由度それぞれが擬似球関節モジュールによって構成されており(\figref{figure:musculoskeletal-structure}の中図), それら関節の角度をポテンショメータによって測定することができる.
  通常筋骨格ヒューマノイドの関節は球関節等により関節角度を直接は測定することが出来ないが, 本構成により, 実験的評価を容易にしている.
  これらの関節のうち, 本研究では主に肩の3自由度と肘の2自由度を中心に扱う.
  MusashiはMusashiLarmの単純な拡張であり, その双腕部をいくつかの実験で用いる.
}%

\subsection{Muscle Configuration} \label{subsec:muscle-configuration}
\switchlanguage%
{%
  Each muscle is actuated by winding Dyneema using a pulley and a brushless DC motor, as shown in the lower center figure of \figref{figure:musculoskeletal-structure}.
  Also, each muscle is folded back by an external pulley, is covered by a spring and a soft foam cover, and an O-ring is inserted in the endpoint of the muscle as a nonlinear elastic element.
  This configuration can realize the nonlinear elasticity of muscles, and we can conduct variable stiffness control and other controls for soft environmental contact.
  MusashiLarm includes a total of 18 muscles, of which 10 muscles are included to move the shoulder and elbow, and 8 muscles are included to move the wrist and fingers, as shown in the right figure of \figref{figure:musculoskeletal-structure}.
}%
{%
  筋構成は\figref{figure:musculoskeletal-structure}の左下図のようになっており, BLDCモータによって化学繊維であるダイニーマをプーリで巻き取ることで腱を駆動する.
  また, 筋は滑車を介して折り返され, その周りを発泡性のカバーとバネが覆うと同時に, 末端には非線形弾性要素としてOリングが挿入されている.
  これにより, 筋の非線形弾性を実現することができ, 可変剛性制御, その他柔らかい接触のための制御を行うことができる.
  また, 全部で筋は18本あり, 肩と肘の5自由度に10本, 前腕から先に8本を用いている(\figref{figure:musculoskeletal-structure}の右図).
}%

\begin{figure}[t]
  \centering
  \includegraphics[width=1.0\columnwidth]{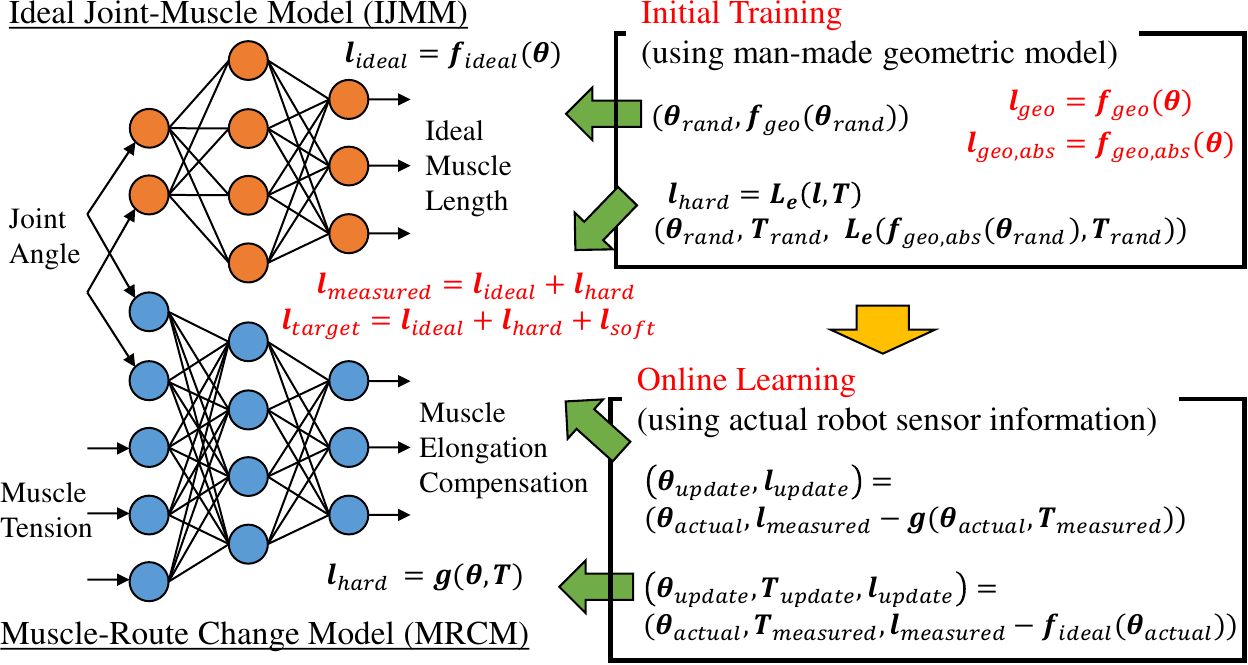}
  \caption{Overview of self-body image acquisition.}
  \label{figure:bodyimage-overview}
  \vspace{-3.0ex}
\end{figure}

\section{Long-time Self-body Image Acquisition} \label{sec:self-body-image}
\subsection{Overview of Self-body Image Acquisition}
\switchlanguage%
{%
  We show the overview of self-body image acquisition in \figref{figure:bodyimage-overview}.
  We define ``the state that can realize intended joint angles'' as the state of having a correct self-body image.
  This definition is special and different from what is called body image or body scheme in neuroscience, etc.

  We express the self-body image by a network structure of which the input is joint angles and muscle tensions and the output is muscle lengths.
  Also, this self-body image has two networks: the first network expresses the ideal relationship between joint angles and muscle lengths in the case that there is no muscle elongation or structure deformation (Ideal Joint-Muscle Mapping, IJMM, $\bm{f}_{ideal}$), and the second network expresses the compensation model of muscle elongation and muscle route changes by muscle tensions (Muscle-Route Change Model, MRCM, $\bm{g}$) as stated below,
  \begin{equation}
    \bm{l} = \bm{f}_{ideal}(\bm{\theta}) + \bm{g}(\bm{\theta}, \bm{T}) \label{eq:model}
  \end{equation}
  where $\bm{l}$ is the measured muscle length, $\bm{T}$ is the measured muscle tension, and $\bm{\theta}$ is the measured joint angle.
  We are able to express the self-body image as one simple neural network, but the scales of the output muscle length of the two networks differ greatly.
  So when we express the self-body image as one simple network, the network cannot learn from the actual robot sensor information well and the learning sometimes proceeds in an unintended direction.
  Therefore, we separate the self-body image into two models: IJMM and MRCM.

  As shown in \figref{figure:bodyimage-overview}, self-body image acquisition has two processes: the first is an initial training of self-body image using the geometric model, and the second is its online learning using the actual robot sensor information.
  The former initializes the weight of the neural network of joint-muscle mapping, and the latter updates it online and constructs the weight for the actual robot.
}%
{%
  自己身体像獲得手法の概要を\figref{figure:bodyimage-overview}に示す.
  本研究では「意図した関節角度を実現できる」ということを自己身体像が正しい状態として定義する.
  ここで述べる定義はいわゆる身体像や身体図式とは異なることに注意していただきたい.

  本研究では, 関節角度と筋張力を入力とし, 筋長を出力とするニューラルネットワークにより自己身体像($\bm{f}$)を表現する.
  この自己身体像には二つの分類があり, 一つ目は筋の伸び等を考えない場合の理想的な関節角度と筋長の関係(理想関節-筋空間マッピング, IJMM, $\bm{f}_{ideal}$), 二つ目は筋張力の影響による筋の伸び・筋経路の変化に対する補償項(筋経路変化補償モデル, MRCM, $\bm{g}$)である.
  \begin{equation}
    \bm{l} = \bm{f}(\bm{\theta}, \bm{T}) = \bm{f}_{ideal}(\bm{\theta}) + \bm{g}(\bm{\theta}, \bm{T}) \label{eq:model}
  \end{equation}
  ここで, $\bm{l}$は測定された筋長, $\bm{T}$は測定された筋張力, $\bm{\theta}$は測定された関節角度を表す.
  これら二つのモデルは合わせて単純な一つのニューラルネットワークとして表現することも可能だが, 出力筋長のスケールが大きく異なり, 合わせた場合には目標出力との差をどちらに逆伝播させるかを選ぶことができず, 学習が進まない, または不都合な方向に学習が進むことが散見されたため, 本研究ではこれら2つのモデルをそれぞれ別のニューラルネットワークとして表現する.

  \figref{figure:bodyimage-overview}に示されるように, 自己身体像獲得には2つの工程があり, 一つ目は幾何モデルを用いた初期学習, 二つ目は実機センサデータを用いたオンライン学習である.
  前者により関節-筋空間マッピングのニューラルネットワークの重みを初期化し, それを実機センサデータから逐次的に更新することで, 実機に適応した重みを形成していく.
}%

\subsection{Comparison of Self-body Images} \label{subsec:self-body-image-difference}
\begin{figure}[t]
  \centering
  \includegraphics[width=0.9\columnwidth]{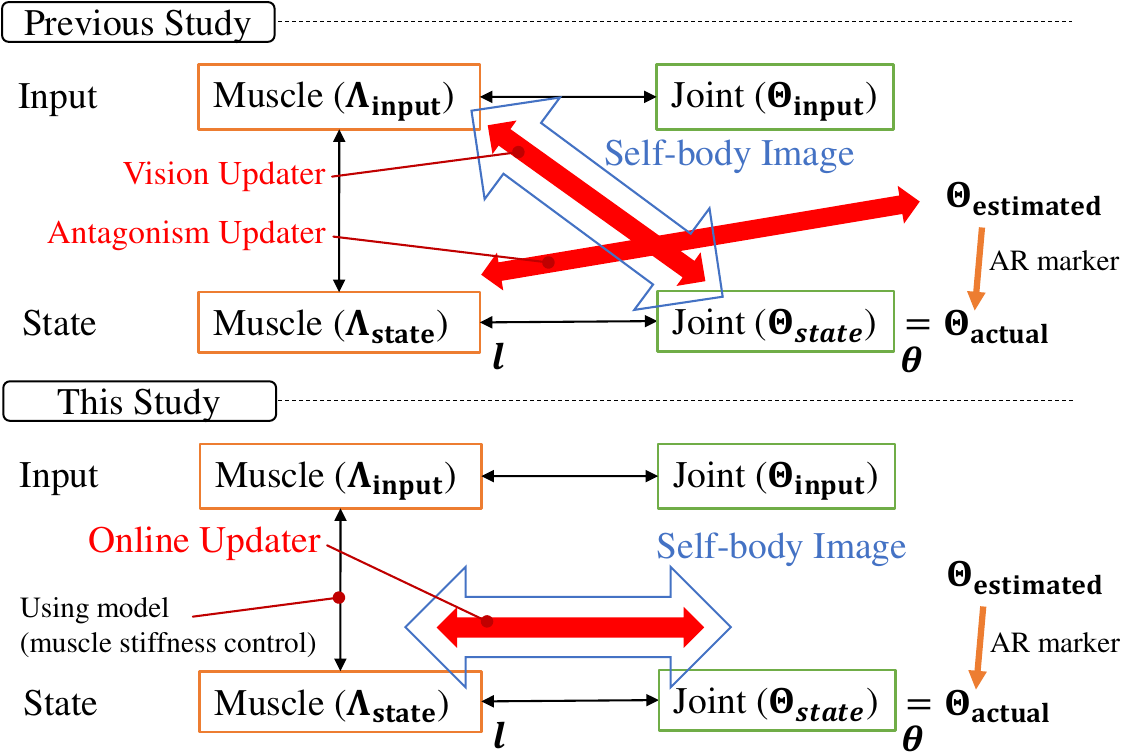}
  \caption{Difference of the network configuration between previous study \cite{kawaharazuka2018bodyimage} and this study.}
  \label{figure:bodyimage-difference}
  \vspace{-3.0ex}
\end{figure}

\switchlanguage%
{%

  The network configuration expressing the self-body image in previous studies \cite{kawaharazuka2018online, kawaharazuka2018bodyimage} and the one in this study are different.
  As shown in \figref{figure:bodyimage-difference}, we define $\bm{\Theta}$ as joint space, $\bm{\Lambda}$ as muscle space, $[\bm{\Theta}\;\bm{\Lambda}]_{input}$ as control input space, and $[\bm{\Theta}\;\bm{\Lambda}]_{state}$ as state space.

  In previous studies \cite{kawaharazuka2018online, kawaharazuka2018bodyimage}, the self-body image is the network between joint state space $\bm{\Theta}_{state}$ and muscle input space $\bm{\Lambda}_{input}$.
  The current joint angle $\bm{\Theta}_{estimated}$ is estimated from the self-body image and $\bm{\Lambda}_{input}$, and the actual current joint angle $\bm{\Theta}_{state} = \bm{\Theta}_{actual}$ is estimated by compensating for $\bm{\Theta}_{estimated}$ using vision.
  There are two updaters: Antagonism Updater and Vision Updater.
  The former updates the antagonism of muscles by learning the relationship between $\bm{\Theta}_{estimated}$ and $\bm{\Lambda}_{state}$, and the latter correctly introduces target muscle lengths which realize the target joint angles by learning the relationship between $\bm{\Theta}_{actual}$ and $\bm{\Lambda}_{input}$, because the information of muscle antagonism is included in $\bm{\Lambda}_{state}$ and we must calculate $\bm{\Lambda}_{input}$.

  In this study, we express self-body image by the relationship between the joint state space $\bm{\Theta}_{state}$ and muscle state space $\bm{\Lambda}_{state}$.
  Because the difference between $\bm{\Lambda}_{input}$ and $\bm{\Lambda}_{state}$ can be calculated through the equation of muscle stiffness control \cite{shirai2011stiffness}, we can calculate $\bm{\Lambda}_{input}$ from $\bm{\Lambda}_{state}$.
  Therefore, there is no need to use Antagonism Updater and Vision Updater, and we need only a simple online updater to learn the relationship between $\bm{\Lambda}_{state}$ and $\bm{\Theta}_{state}$.
  Thus, the networks in previous studies express the software and hardware elasticity, but the one in this study expresses only the hardware elasticity and handles the software elasticity separately.

  This mechanism can not only integrate 2 updaters, but also be generally applied to torque control by muscle tension, etc., because the network uses only the value of state space.
  In the following sections, we introduce the detailed system of this study.
}%
{%
  \cite{kawaharazuka2018online, kawaharazuka2018bodyimage}における自己身体像と本研究における自己身体像を表すネットワークの構成は異なる.
  \figref{figure:bodyimage-difference}に示すように, 関節空間$\bm{\Theta}$と筋空間$\bm{\Lambda}$, そして制御入力$\cdot_{input}$と状態$\cdot_{state}$を表す空間を定義する.

  先行研究\cite{kawaharazuka2018online, kawaharazuka2018bodyimage}では, 自己身体像は関節状態$\bm{\Theta}_{state}$と筋制御入力$\bm{\Lambda}_{input}$を対応付けたネットワークとなっている.
  筋制御入力$\bm{\Lambda}_{input}$から現在関節状態$\bm{\Theta}_{estimated}$を推定し, これを視覚により補正することで実機関節状態$\bm{\Theta}_{state} = \bm{\Theta}_{actual}$を推定する.
  オンライン学習にはAntagonism UpdaterとVision Updaterが存在し, これらはそれぞれ, $\bm{\Theta}_{estimated}$と$\bm{\Lambda}_{state}$の関係を学習することで拮抗関係を修正し, $\bm{\Theta}_{actual}$と$\bm{\Lambda}_{input}$の関係を学習することである関節状態を実現する筋制御入力を導出できるようにしていた.
  拮抗関係の情報は$\bm{\Lambda}_{state}$に含まれ, 最終的に導出するべきは$\bm{\Lambda}_{input}$であるため, このような手法となっている.

  本研究では, 自己身体像を関節状態$\bm{\Theta}_{state}$と筋状態$\bm{\Lambda}_{state}$を対応付けたネットワークとして表す.
  $\bm{\Lambda}_{input}$と$\bm{\Lambda}_{state}$の違いは筋剛性制御\cite{shirai2011stiffness}の計算式を通して求まるため, これを逆に計算することで$\bm{\Lambda}_{input}$を求める.
  そのため, Antagonism UpdaterとVision Updaterは不要となり, $\bm{\Lambda}_{state}$と$\bm{\Theta}_{state}$の関係を学習する一つの更新則のみが必要となる.
  つまり, \cite{kawaharazuka2018bodyimage}におけるMRCMはソフトウェアばね, ハードウェアばねの影響を統合して加味していたのに対して, 本研究ではハードウェアばねによる影響のみを含め, ソフトウェアばねの影響をモデルから求める.

  これは, 2種類の更新則をシンプルに1つにまとめられるだけでなく, 状態空間で全てが解決するため, 汎用的に筋張力によるトルク制御等にも応用することができる.
  以降で本研究の具体的な手法について述べる.
}%

\subsection{Initial Training of Self-body Image Using a Geometric Model} \label{subsec:initial-training}
\switchlanguage%
{%
  First, we train IJMM using the geometric model which linearly expresses muscle routes by the start point, relay points, and end point (the right figure of \figref{figure:musculoskeletal-structure}).
  We move joints of the geometric model variously in the range of the joint angle limit, calculate relative muscle lengths from the initial posture (all joint angles are $\bm{0}$ as shown in the right figure of \figref{figure:musculoskeletal-structure}), and train IJMM using these pair data of joint angles $\bm{\theta}_{rand}$ and muscle lengths $\bm{f}_{geo}(\bm{\theta}_{rand})$ (the upper right figure of \figref{figure:bodyimage-overview}).
  Second, regarding MRCM, we approximate the relationship between muscle tension and muscle elongation $\bm{L}_{e}$ with an exponential function using a test sample of one muscle.
  Then we make a dataset of joint angles $\bm{\theta}_{rand}$, muscle tensions $\bm{T}_{rand}$, and compensating value of muscle lengths $-\bm{L}_{e}(\bm{f}_{geo, abs}(\bm{\theta}_{rand}), \bm{T}_{rand})$ considering the elongation of the Dyneema in proportion to the absolute muscle lengths, and train MRCM (the lower right of \figref{figure:bodyimage-overview}).
  In these procedures, $\bm{f}_{geo}$ calculates relative muscle lengths from the initial posture, and $\bm{f}_{geo, abs}$ calculates absolute muscle lengths, from the geometric model.
  The current IJMM approximates the muscle routes along bone structures linearly, and the current MRCM cannot consider influences of muscle interferences, the soft foam cover, structure deformation, etc., so we need to update the self-body image using the actual robot sensor information.
}%
{%
  まず, IJMMに関しては, 筋の取り付け点を起始点・中継点・終止点で表し直線で結んだ幾何モデルを用いて学習させる.
  それぞれの関節を可動域の中で様々に動かし, 初期姿勢(関節角度が全て0)からの相対筋長を幾何モデルから計算し, この関節角度と相対筋長のペアを用いてIJMMを学習させる(\figref{figure:bodyimage-overview}の右上).
  また, MRCMに関しては, 筋単体のテストベッドにより筋張力と筋の伸びの関係を関数フィッティングにより求め, 筋長の全体長さに比例するダイニーマの伸びを考慮して関節角度, 筋張力, 筋長のデータセットを作成し, MRCMを学習させる(\figref{figure:bodyimage-overview}の右下).
  IJMMは本来骨格を沿うような経路を取る筋を直線で近似しており, MRCMは筋同士の干渉や筋を覆う発泡性カバーの影響等を考慮できておらず, これら初期学習により得られた自己身体像を実機センサデータによって更新していく必要がある.
}%

\subsection{Online Learning of Self-body Image Using the Actual Robot} \label{subsec:online-learning}
\switchlanguage%
{%
  First, the data used for the online learning of self-body image is shown as below,
  \begin{align}
    \bm{\theta}_{vision} &= IK(\bm{\theta}_{initial}=\bm{\theta}_{est}, \bm{P}_{target}=\bm{P}_{vision}) \\
    \bm{\theta}_{actual} &= \bm{\theta}_{potentio}\;\; \textrm{or}\;\; \bm{\theta}_{vision} \\
    (\bm{\theta}_{update}&, \bm{l}_{update}) = (\bm{\theta}_{actual}, \bm{l}_{m}-\bm{g}(\bm{\theta}_{actual}, \bm{T}_{m})) \label{eq:ijmm-learn}\\
    (\bm{\theta}_{update}&, \bm{T}_{update}, \bm{l}_{update}) = \nonumber\\
                         &\quad(\bm{\theta}_{actual}, \bm{T}_{m}, \bm{l}_{m}-\bm{f}_{ideal}(\bm{\theta}_{actual})) \label{eq:mrcm-learn}
  \end{align}
  where $\bm{l}_{m}(\bm{l}_{measured})$ is muscle lengths measured by encoders in muscle motors, $\bm{T}_{m}(\bm{T}_{measured})$ is muscle tensions measured by loadcells in muscle modules, $\bm{\theta}_{vision}$ is the estimated actual joint angles using vision implemented in \cite{kawaharazuka2018online} (solve inverse kinematics (IK) by setting the initial value $\bm{\theta}_{initial}$ as the estimated joint angles $\bm{\theta}_{est}$ and the target value $\bm{P}_{target}$ as the position of AR marker attached to the end effector $\bm{P}_{vision}$), $\bm{\theta}_{potentio}$ is the joint angles of potentiometers in the joint modules of \cite{kawaharazuka2018musashilarm-en}, and $\{\bm{\theta, T, l}\}_{update}$ is the data used for the online learning.
  We use $\bm{\theta}_{potentio}$ as $\bm{\theta}_{actual}$, because each joint has potentiometers, which is one feature of Musashi used in this study.
  However, when the humanoid has no joint angle sensors like Kengoro \cite{asano2016kengoro}, the online learning can be done by using $\bm{\theta}_{vision}$.
  Because the self-body image in this study describes the relationship of measured joint angles, muscle tensions, and muscle lengths, the online updater in \equref{eq:ijmm-learn} updates IJMM, which is the ideal relationship between joint angles and muscle lengths, by removing the influence of MRCM, and the online updater in \equref{eq:mrcm-learn} updates MRCM, which is the compensation value of muscle elongation and muscle route changes by muscle tensions, by removing the influence of IJMM.
  Also, this updater generates data for online learning at 2 Hz when the current joint angles or muscle lengths deviate from the previously learned data.
}%
{%
  まず, 自己身体像のオンライン学習で用いるデータは以下のようになる.
  \begin{align}
    \bm{\theta}_{vision} &= IK(\bm{\theta}_{initial}=\bm{\theta}_{est}, \bm{P}_{target}=\bm{P}_{vision}) \\
    \bm{\theta}_{actual} &= \bm{\theta}_{potentio}\;\; or\;\; \bm{\theta}_{vision} \\
    (\bm{\theta}_{update}&, \bm{l}_{update}) = (\bm{\theta}_{actual}, \bm{l}_{m}-\bm{g}(\bm{\theta}_{actual}, \bm{T}_{m})) \label{eq:ijmm-learn}\\
    (\bm{\theta}_{update}&, \bm{T}_{update}, \bm{l}_{update}) = \nonumber\\
                         &\quad(\bm{\theta}_{actual}, \bm{T}_{m}, \bm{l}_{m}-\bm{f}_{ideal}(\bm{\theta}_{actual})) \label{eq:mrcm-learn}
  \end{align}
  となる.
  ここで, $\bm{l}_{m}(\bm{l}_{measured})$は筋モータのエンコーダから測定された筋長, $\bm{T}_{m}(\bm{T}_{measured})$は筋モジュールに搭載された筋張力センサの値, $\bm{\theta}_{vision}$は\cite{kawaharazuka2018online}で実装された関節角度推定とARマーカを用いた視覚による実機関節角度推定値(初期値$\bm{\theta}_{initial}$を関節角度推定値$\bm{\theta}_{est}$, ターゲット$\bm{P}_{target}$を手先のARマーカの位置$\bm{P}_{vision}$として逆運動学(IK)を解く), $\bm{\theta}_{potentio}$は本研究で用いる筋骨格上肢に特有の各関節に入ったポテンショメータの値, $\{\bm{\theta, T, l}\}_{update}$は学習に用いるデータである.
  本研究で用いる筋骨格上肢の特徴として, 各関節にポテンショメータが入っているため, 本研究では$\bm{\theta}_{actual}$として$\bm{\theta}_{potentio}$を用いるが, 腱悟郎\cite{asano2016kengoro}のように関節角度が直接測定できない筋骨格ヒューマノイドでも, 視覚さえあれば$\bm{\theta}_{vision}$を用いて同じように逐次学習を行うことができる.
  本研究で考える自己身体像は, 測定された関節角度, 筋張力, 筋長の関係を記述するため, \equref{eq:ijmm-learn}ではMRCMの影響を消去することで関節角度と理想的な筋長の関係であるIJMMを学習し, \equref{eq:mrcm-learn}においてはIJMMの影響を消去することで筋張力による筋の伸びや筋経路変化を補償するための項であるMRCMを学習する.
  また, これは関節角度か筋張力が前回の学習の際のデータからある閾値以上離れた場合に学習データを作成する.
}%

\begin{figure}[t]
  \centering
  \includegraphics[width=1.0\columnwidth]{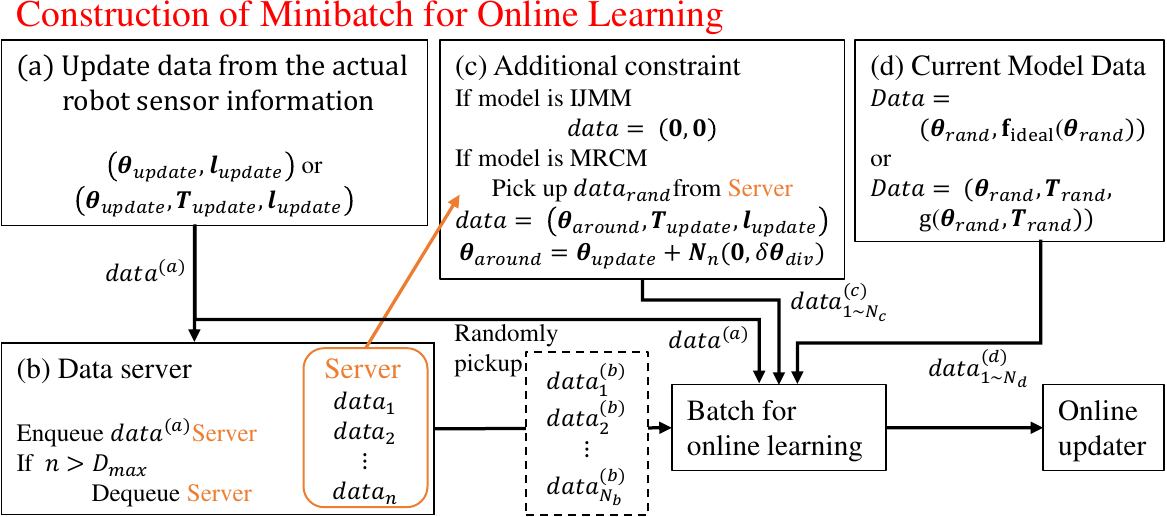}
  \caption{Data accumulation and augmentation of the actual robot sensor information for generation of minibatch for the stable online learning of self-body image.}
  \label{figure:data-flow}
  \vspace{-3.0ex}
\end{figure}

\switchlanguage%
{%
  Next, we show how to accumulate and augment the actual robot sensor information for the generation of minibatch for online learning in \figref{figure:data-flow}.
  In procedure (a), the data from the actual robot sensor information for the online learning is extracted as shown in \equref{eq:ijmm-learn} -- \equref{eq:mrcm-learn}.
  After that, the data obtained from (a) is accumulated in the data server of (b).
  Procedure (c) generates the data which adds the necessary limitation to the network structure of IJMM and MRCM.
  In the case of IJMM, because the relative muscle lengths are $\bm{0}$ when all joint angles are $\bm{0}$, (c) generates the data $(\bm{0}, \bm{0})$.
  In the case of MRCM, because the nonlinear elastic relationship between muscle lengths and muscle tensions does not change greatly according to joint angles, (c) generates the data $(\bm{\theta}_{around}, \bm{T}_{update}, \bm{l}_{update})$.
  $\{\bm{\theta}, \bm{T}, \bm{l}\}_{update}$ is the actual robot sensor data extracted from the data server of (b), and $\bm{\theta}_{around}$ is obtained by adding random values following a normal distribution with an average of $0$ and a dispersion of $\delta\bm{\theta}_{div}$ to $\bm{\theta}_{update}$.
  Procedure (d) generates data by inputting $\bm{\theta}, \bm{T}$ randomly into the current model and obtaining the output.
  The data is restricted by the fact that the data space other than obtained sensor data should not change from the current model.
  Finally, we extract one piece of data from (a), $N_{b}, N_{c}, N_{d}$ numbers of data from (b), (c), and (d), respectively, and generate a minibatch for online learning by compiling them together.
  In this study, we set $N_{b}=10, N_{c}=5, N_{d}=5$.
}%
{%
  次に, データをどう蓄積・拡張してバッチとして用いるかを\figref{figure:data-flow}に示す.
  まず\figref{figure:data-flow}の(a)は\equref{eq:ijmm-learn}--\equref{eq:mrcm-learn}に示されるような, 実機センサデータから学習データを抽出する部分である.
  その後, (a)で得られたデータを(b)のデータサーバに蓄積していく.
  (c)ではIJMM, MRCMの構造に合わせて必要な制限を加えるデータを作成する.
  IJMMの場合は, 関節角度が全て$\bm{0}$のときに, 相対筋長は$\bm{0}$であるという制限を加え, $(\bm{0}, \bm{0})$というデータを作成する.
  MRCMの場合は, 関節角度によって大きくは筋長と筋張力の非線形弾性関係は変わらないという制限を加え, $(\bm{\theta}_{around}, \bm{T}_{update}, \bm{l}_{update})$というデータを作成する.
  ここで, $\bm{\theta}_{update}, \bm{T}_{update}, \bm{l}_{update}$は(b)のデータサーバからランダムに取り出した実機センサデータであり, $\bm{\theta}_{around}$は$\bm{\theta}_{update}$に平均$0$, 分散$\delta\bm{\theta}_{div}$の正規分布に従う乱数を加えた値である.
  (d)では現状のモデルにランダムに$\bm{\theta}, \bm{T}$を与え出力を得て, それらをデータセットとして用いる.
  これは, センサデータを得られた空間以外の空間は現状のモデルから学習によって大きく変わるべきではない, という制約によるデータセットである.
  最終的に, (a)のデータ, (b)(c)(d)からそれぞれ$N_{b}, N_{c}, N_{d}$個得られたデータの集合を学習用のバッチとし, オンライン学習を行う.
  本研究では$N_{b}=10, N_{c}=5, N_{d}=5$としている.
}%

\subsection{Safety Mechanism Considering Muscle Tension and Temperature} \label{subsec:stable-learning}
\switchlanguage%
{%
  In order to move the musculoskeletal humanoid for a long time while acquiring self-body image, it is important to prevent damage to the muscle motor by unintended high muscle tension and burnout of the muscle motor by high motor temperature.
  Therefore, we adjust the target muscle length to suppress the unintended high muscle tension and temperature as shown below,
  \begin{align}
    \delta{l} =& K_{T}\textrm{max}(T-T_{lim}, 0)+K_{C}\textrm{max}(C-C_{lim}, 0)\nonumber\\
    \delta{l}_{t+1} =& \delta{l}_{t} + \textrm{max}(-\delta{l}_{lim}, \textrm{min}(\delta{l}_{lim}, \delta{l}-\delta{l}_{t}))\nonumber\\
    l_{target} =& l_{target}+\delta{l}_{t+1} \label{eq:safety-mechanism}
  \end{align}
  where $l_{target}$ is the target muscle length, $\delta{l}$ is the ideal relative change of $l_{target}$, $T, C$ are the current muscle tension and temperature, $K_{T}, K_{C}$ are the gains that inhibit $T$ and $C$, respectively, $T_{lim}, C_{lim}$ are the threshold values for the inhibition of the rise in $T$ and $C$, $\delta{l}_{lim}$ is the limitation threshold of the change in relative muscle length for the motor not to vibrate, and $\delta{l}_{t+1}$ is the regulated relative change of muscle length which is sent at the current step.
  In this study, we set $K_{T}=1.0$ [mm/N], $K_{C}=1.0$ [mm/${}^\circ\textrm{C}$], $T_{lim}=200$ [N], $C_{lim}=60$ [${}^\circ\textrm{C}$], $\delta{l}_{lim}=0.01$ [mm], and this safety mechanism runs every 8 msec.
  This safety mechanism can inhibit high muscle tension, and cope with the case in which muscle tension is not high but the muscle temperature rises gradually, though the tracking ability of joint angles deteriorates to a certain degree.
}%
{%
  長期的に自己身体像を獲得しつつ動かすためには, 意図しない筋張力によりロボットが破損したり熱により筋モータが焼損したりするのを防ぐことが重要となる.
  そこで, 以下のように筋張力と温度の上昇に従って指令する筋長を変更し, 意図しない筋張力と温度の上昇を抑える手法を考案した.
  \begin{align}
    \delta{\bm{l}} =& K_{T}\textrm{max}(\bm{T}-\bm{T}_{lim}, \bm{0})+K_{C}\textrm{max}(\bm{C}-\bm{C}_{lim}, \bm{0})\nonumber\\
    \delta{\bm{l}}_{t+1} =& \delta{\bm{l}}_{t} + \textrm{max}(-\delta{\bm{l}}_{lim}, \textrm{min}(\delta{\bm{l}}_{lim}, \delta{\bm{l}}-\delta{\bm{l}}_{t}))\nonumber\\
    \bm{l}_{target} =& \bm{l}_{target}+\delta{\bm{l}}_{t+1} \label{eq:safety-mechanism}
  \end{align}
  $\delta{\bm{l}}$は現在の指令筋長$\bm{l}_{target}$に足しこむべき差分筋長の理想値, $K_{T}, K_{C}$はそれぞれ筋張力と筋温度の上昇を抑えるためのゲイン, $\bm{T}, \bm{C}$は現在の筋張力と筋温度, $\bm{T}_{lim}, \bm{C}_{lim}$はそれぞれ筋張力と筋温度の上昇を抑え始める値の閾値, $\delta{\bm{l}}_{t+1}$は次のタイムステップに送る差分筋長, $\delta{\bm{l}}_{lim}$は差分筋長の理想値に振動せず徐々に近づけるための閾値を表す.
  本研究では, $K_{T}=1.0$ [mm/N], $K_{C}=1.0$ [mm/${}^\circ\textrm{C}$], $\bm{T}_{lim}=200$ [N], $\bm{C}_{lim}=60$ [${}^\circ\textrm{C}$], $\delta{\bm{l}}_{lim}=0.01$ [mm]に設定し, 8 [msec]で動作させている.
  この安全機構により, 大きな筋張力を抑制し, また, 筋張力は大きくないが徐々に温度が上昇していく場合にも対応することができる.
}%

\section{Position, Torque, and Variable Stiffness Control Using Self-body Image} \label{sec:variable-stiffness}

\subsection{Position Control}
\switchlanguage%
{%
  First, we will explain position control using the self-body image.
  In this study, we move Musashi using muscle stiffness control \cite{shirai2011stiffness} as shown below,
  \begin{align}
    \bm{T}_{target} = \bm{T}_{bias} + \textrm{max}(\bm{0}, K_{stiff}(\bm{l}-\bm{l}_{target})) \label{eq:muscle-stiffness-control}
  \end{align}
  where $\bm{T}_{target}$ is the target muscle tension, $\bm{T}_{bias}$ is the bias term of the muscle stiffness control, and $K_{stiff}$ is the software muscle stiffness.
  In order to realize the intended joint angles, we need to decide $\bm{l}_{target}$ in \equref{eq:muscle-stiffness-control}, and this is done as shown below,
  \begin{align}
    \bm{l}_{soft}(\bm{T}) &= -(\bm{T} - \bm{T}_{bias})/K_{stiff} \\
    \bm{l}_{target} &= \bm{f}(\bm{\theta}_{target}, \bm{T}_{const})+\bm{l}_{soft}(\bm{T}_{const}) \label{eq:move-first}\\
    \bm{l}_{target} &= \bm{f}(\bm{\theta}_{target}, \bm{T}_{measured})+\bm{l}_{soft}(\bm{T}_{measured}) \label{eq:move-second}
  \end{align}
  where $\bm{f}(\bm{\theta}, \bm{T})$ is the self-body image, $\bm{T}_{const}$ is constant muscle tension, and $\bm{l}_{soft}$ is the compensation values of software muscle elasticity from $\bm{l}_{target}$ by muscle stiffness control.
  In \equref{eq:move-first}, we move the musculoskeletal humanoid using the target joint angles and constant muscle tension.
  However, although this movement can realize the target joint angles to a certain degree, the robot cannot move to the target posture completely because the target muscle tension is impossible to realize.
  Then, in \equref{eq:move-second}, when we set the current measured muscle tensions to the target muscle tensions, they become the necessary muscle tensions to approximately realize the target joint angles.
  Ideally the robot can completely realize the target joint angles by continuing the feedback of the current measured muscle tensions, but in actuality, there is an error between the self-body image and the actual robot, so the current muscle tensions can diverge or converge to minimum muscle tensions $\bm{T}_{bias}$, which is not practical, if we continue the feedback of \equref{eq:move-second}.
  Also, although there are many combinations of target muscle tensions which can realize the target joint angles due to the redundant muscle arrangements, by setting the $\bm{T}_{const}$ to the minimum muscle tension $\bm{T}_{bias}$, the robot can realize the target joint angles by minimum muscle tensions.
}%
{%
  まずは, 位置制御について説明する.
  本研究では筋骨格ヒューマノイドを以下のような筋剛性制御\cite{shirai2011stiffness}により動作させる.
  \begin{align}
    \bm{T}_{target} = \bm{T}_{bias} + \textrm{max}(\bm{0}, K_{stiff}(\bm{l}-\bm{l}_{target})) \label{eq:muscle-stiffness-control}
  \end{align}
  ここで, $\bm{T}_{target}$は指令筋張力, $\bm{T}_{bias}$は筋剛性制御のバイアス項, $K_{stiff}$は筋剛性制御の筋剛性係数である.
  ある関節角度を実現したい場合, \equref{eq:muscle-stiffness-control}における$\bm{l}_{target}$を決める必要があり, それは以下のように行う.
  \begin{align}
    \bm{l}_{soft}(\bm{T}) &= -(\bm{T} - \bm{T}_{bias})/K_{stiff} \\
    \bm{l}_{target} &= \bm{f}(\bm{\theta}_{target}, \bm{T}_{const})+\bm{l}_{soft}(\bm{T}_{const}) \label{eq:move-first}\\
    \bm{l}_{target} &= \bm{f}(\bm{\theta}_{target}, \bm{T}_{measured})+\bm{l}_{soft}(\bm{T}_{measured}) \label{eq:move-second}
  \end{align}
  ここで, $\bm{f}(\bm{\theta}, \bm{T})$は自己身体像, $\bm{T}_{const}$はある一定の筋張力, $\bm{l}_{soft}$は筋剛性制御によるソフトウェアでの$\bm{l}_{target}$からの筋の伸びを補償するための項である.
  \equref{eq:move-first}では指令関節角度と, ある適当な筋張力から筋骨格ヒューマノイドを動作させるが, このとき実機関節角度はある程度指令値まで追従するものの, 実現不可能な筋張力を送っている可能性があるため, 実際に指令した関節角度が正しく実現できるとは限らない.
  そこで, \equref{eq:move-second}で現在の筋張力を指令筋張力とすることで, 指令関節角度付近で実際に必要な筋張力が指令筋張力となるため, より正しく指令関節角度を実現することができることとなる.
  理想的には, 筋張力をフィードバックし続ければ正しい指令関節角度を実現できるが, 実際には自己身体像と実機の間に誤差が存在し, そのためフィードバックし続けると, 筋張力が発散または$\bm{T}_{bias}$に収束し得るため, 現実的ではない.
  またこのとき, 冗長な筋配置のため指令関節角度を実現することのできる指令筋張力は多数存在し, $\bm{T}_{const}$を最小の$\bm{T}_{bias}$にすることで, 最小限の筋張力で指令関節角度を実現することができる.
}%

\subsection{Torque Control}
\switchlanguage%
{%
  Next, by using this self-body image, we can conduct torque control of musculoskeletal structures.
  The basic method is the joint torque control using muscle tension \cite{kawamura2016jointspace}.
  By the acquisition of self-body image, the torque control \cite{kawamura2016jointspace} becomes better, because a more accurate muscle Jacobian than the one of the geometric model can be obtained by differentiating the self-body image.
  Also, the acquisition of self-body image makes the joint angle estimation more correct \cite{kawaharazuka2018online, kawaharazuka2018bodyimage}, and the actual joint angles can follow the target joint angles more accurately.
}%
{%
  次に, この自己身体像を用いることで, トルク制御を行うことができる.
  ベースとなる手法は\cite{kawamura2016jointspace}における筋張力によるトルク制御である.
  自己身体像の逐次的獲得により, この自己身体像を微分することで幾何モデルよりも正しい筋長ヤコビアンが得られるため, \cite{kawamura2016jointspace}のような筋張力制御をより正確に実行することができる.
  また, 自己身体像の逐次的獲得により, 筋長・筋張力の変化からの関節角度推定もより正しくなるため, 関節角度の指令値追従度も向上する.
}%

\subsection{Variable Stiffness Control} \label{subsec:variable-stiffness}

\begin{figure}[t]
  \centering
  \includegraphics[width=1.0\columnwidth]{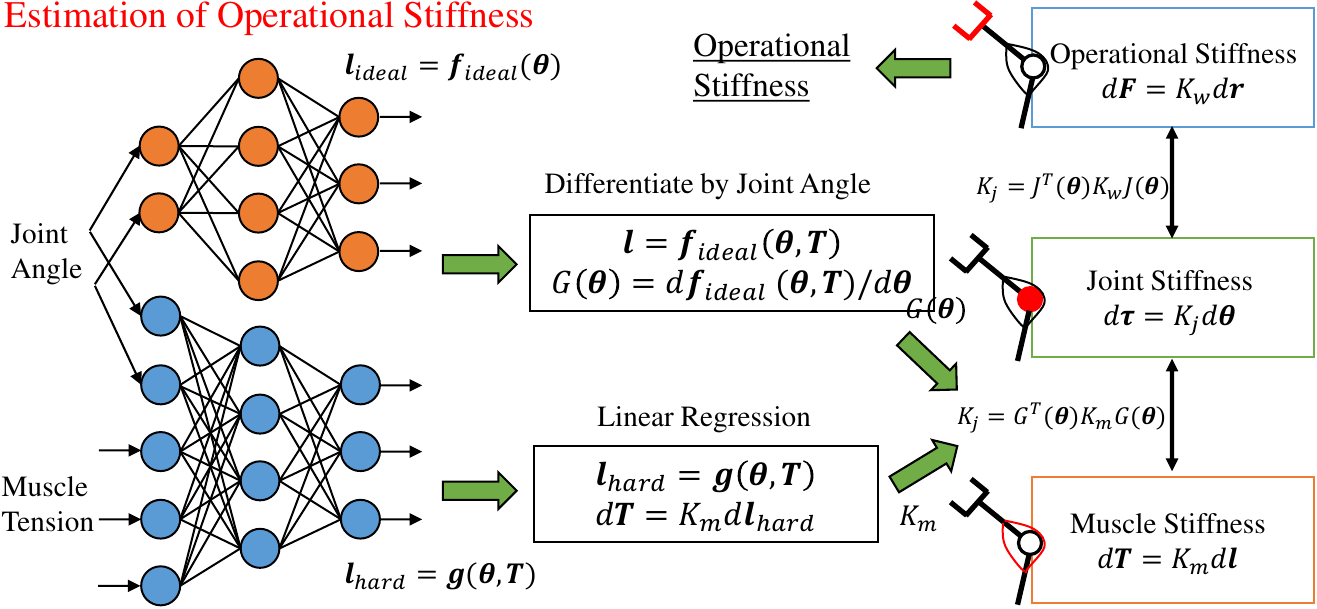}
  \caption{Estimation of operational hardware stiffness using self-body image.}
  \label{figure:variable-stiffness-estimation}
  \vspace{-3.0ex}
\end{figure}

\subsubsection{Estimation of Self-body Stiffness}
\switchlanguage%
{%
  First, we estimate operational hardware stiffness as shown in \figref{figure:variable-stiffness-estimation}.
  In order to estimate the operational stiffness, joint Jacobian $J$, muscle Jacobian $G$, and muscle stiffness $K_m$ are necessary.
  First, we can obtain the joint Jacobian from the geometric model, as with ordinary axis-driven humanoids.
  Second, muscle Jacobian can be obtained by differentiating the IJMM of the self-body image by the joint angles.
  Third, muscle stiffness can be obtained using linear regression between the change in muscle tensions and the change in output of the MRCM.
  Therefore, the operational stiffness can be estimated by multiplying muscle stiffness by muscle Jacobian and joint Jacobian.
}%
{%
  自己身体像を用いた作業空間剛性の導出の概要を\figref{figure:variable-stiffness-estimation}に示す.
  \figref{figure:variable-stiffness-estimation}の右図にあるように, 作業空間剛性を導出するためには, 関節ヤコビアン$J$, 筋長ヤコビアン$G$, 筋剛性$K_m$, の3つが必要となる.
  まず, 関節ヤコビアンは通常の軸関節型ロボット同様, 幾何モデルから直接得ることが可能である.
  人体模倣型の筋骨格ヒューマノイドの場合, 球関節等のリンク長が変わったり回転中心が変わったりする関節が存在する可能性があるが, 本研究で用いる筋骨格上肢にそのような構造はないため, 幾何モデルを用いる.
  次に, 筋長ヤコビアンは自己身体像におけるIJMMを関節角度で微分することで求めることができる.
  最後に, 筋剛性は筋張力変化とMRCMの出力変化の線形回帰によって求めることができる.
  よって, 筋剛性に筋長ヤコビアンと関節ヤコビアンをかけることで, 作業空間剛性を導出することができる.
}%

\begin{figure}[t]
  \centering
  \includegraphics[width=1.0\columnwidth]{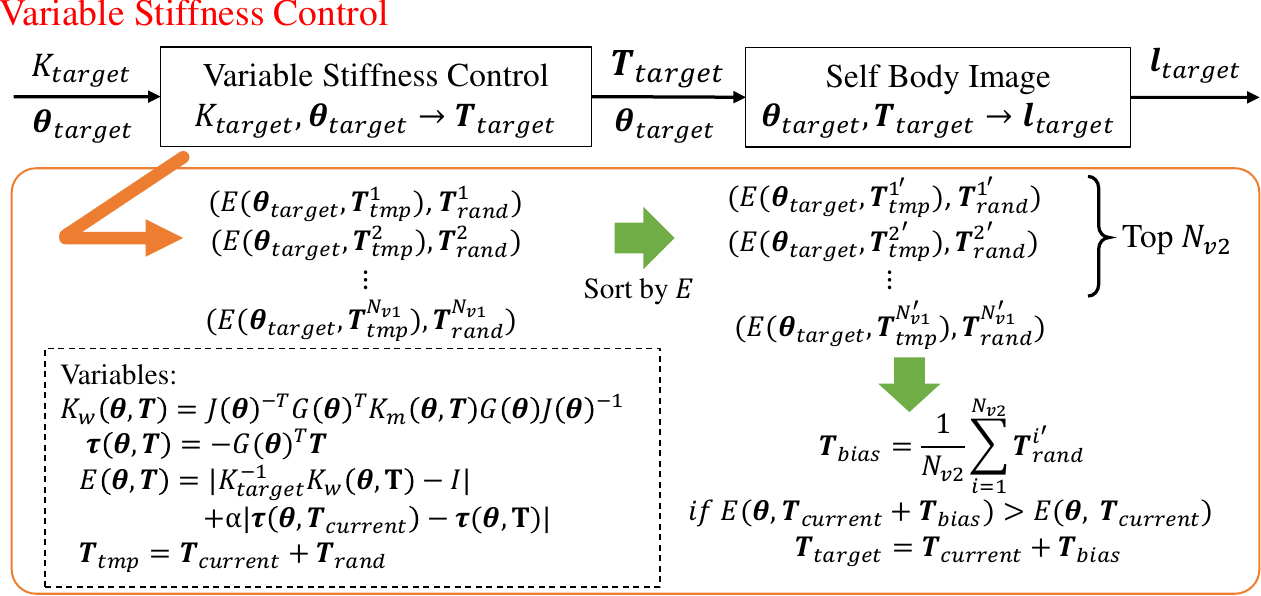}
  \caption{Control of operational hardware stiffness using self-body image.}
  \label{figure:variable-stiffness-control}
  \vspace{-3.0ex}
\end{figure}

\subsubsection{Control of Self-body Stiffness}
\switchlanguage%
{%
  The basic system of variable stiffness control is shown in the upper figure of \figref{figure:variable-stiffness-control}: we set the target joint angles and target operational stiffness, calculate the target muscle tension using the method we will propose in this subsection, calculate the target muscle lengths from self-body image by inputting the calculated muscle tensions, and send them to the actual robot.
  The details of variable stiffness control is shown in the lower figure of \figref{figure:variable-stiffness-control}.
  First, we make $\bm{T}_{tmp}$ by adding $\bm{T}_{rand}$, which is a random value within a certain range ($-20\sim20$ [N] in this study), to the current muscle tensions $\bm{T}_{current}$, and accumulate $N_{v1}$ data pairs of $\bm{T}_{rand}$ and evaluation value $E(\bm{\theta}_{target}, \bm{T}_{tmp})$ based on the equation as shown below,
  \begin{align}
    \bm{T}_{tmp}=&\bm{T}_{current}+\bm{T}_{rand} \nonumber\\
    K_w(\bm{\theta}, \bm{T})=&J(\bm{\theta})^{-T}G(\bm{\theta})^{T}K_m(\bm{\theta}, \bm{T})G(\bm{\theta})J(\bm{\theta})^{-1} \nonumber\\
    \bm{\tau}(\bm{\theta}, \bm{T})=&-G(\bm{\theta})^{T}\bm{T} \nonumber\\
    E(\bm{\theta}, \bm{T})=&|K^{-1}_{target}K_w(\bm{\theta}, \bm{T})-I| \nonumber \\
                           &+\alpha|\bm{\tau}(\bm{\theta}, \bm{T}_{current})-\bm{\tau}(\bm{\theta}, \bm{T})|
  \end{align}
  where $K_w(\bm{\theta}, \bm{T})$ is the calculated operational stiffness, $K_{target}$ is the target operational stiffness, $J(\bm{\theta})$ is the joint Jacobian, $G(\bm{\theta})$ is the muscle Jacobian, $\bm{K}_{m}$ is the muscle stiffness, $\bm{\tau}(\bm{\theta}, \bm{T})$ is the joint torque, $\alpha$ is a weight constant, and $|\cdot|$ expresses L2 norm.
  The evaluation value $E(\bm{\theta}, \bm{T})$ sums up the values that express how the calculated operational stiffness is close to the target operational stiffness when $\bm{T}=\bm{T}_{target}$, and how the calculated joint torques are close to the current joint torques when $\bm{T}=\bm{T}_{current}$, by the weight of $1:\alpha$.
  We sort the accumulated data in ascending order of the value $E$, and set $\bm{T}_{bias}$ as the average $\bm{T}_{rand}$ of $N_{v2}$ ($N_{v1}>N_{v2}$) data from the top.
  Then, we replace the $\bm{T}_{target}$ by $\bm{T}_{current}+\bm{T}_{bias}$, when the value $E(\bm{\theta}_{target}, \bm{T}_{current}+\bm{T}_{bias})$ is smaller than $E(\bm{\theta}_{target}, \bm{T}_{current})$.
  We repeat this step $N_{v3}$ times, input the final calculated value of $\bm{T}_{target}$ into the self-body image, and obtain the target muscle length.
  When $N_{v2}=1$, this method is a hill climbing method, and when $N_{v2}>1$, the stiffness search becomes more stable.
  In this study, we set $\alpha=0.02, N_{v1}=10, N_{v2}=2$, and $N_{v3}=50$.

  Although we change the operational stiffness in this section, this method can also be applied to change the joint stiffness.
}%
{%
  自己身体像を用いた作業空間剛性制御の概要を\figref{figure:variable-stiffness-control}に示す.
  基本的な構成としては, \figref{figure:variable-stiffness-control}の上図にあるように, ターゲットとなる関節角度と作業空間剛性を入力とし, 本節の手法によって計算し得られた筋張力を入力として自己身体像から指令筋長を算出, ロボットに送る.
  作業空間剛性制御の詳細は\figref{figure:variable-stiffness-control}の下図のようになっている.
  まず, 現状の筋張力$\bm{T}_{current}$にある範囲内の(本研究では-20 [N]から20 [N])のランダムな$\bm{T}_{rand}$を加え$\bm{T}_{tmp}$とし, それを以下の式に基づいて評価値$E(\bm{\theta}_{target}, \bm{T}_{tmp})$を求めたデータを集積する.
  \begin{align}
    \bm{T}_{tmp}=&\bm{T}_{current}+\bm{T}_{rand} \nonumber\\
    K_w(\bm{\theta}, \bm{T})=&J(\bm{\theta})^{-T}G(\bm{\theta})^{T}K_m(\bm{\theta}, \bm{T})G(\bm{\theta})J(\bm{\theta})^{-1} \nonumber\\
    \bm{\tau}(\bm{\theta}, \bm{T})=&-G(\bm{\theta})^{T}\bm{T} \nonumber\\
    E(\bm{\theta}, \bm{T})=&|K^{-1}_{target}K_w(\bm{\theta}, \bm{T})-I| \nonumber \\
                           &+\alpha|\bm{\tau}(\bm{\theta}, \bm{T}_{current})-\bm{\tau}(\bm{\theta}, \bm{T})|
  \end{align}
  ここで, $K_w(\bm{\theta}, \bm{T})$は作業空間剛性, $K_{target}$は作業空間剛性の指令値, $J(\bm{\theta})$は関節ヤコビアン, $G(\bm{\theta})$は筋長ヤコビアン, $\bm{K}_{m}$は筋剛性, $\bm{\tau}(\bm{\theta}, \bm{T})$は関節トルク, $\alpha$は重みの定数, $|\cdot|$はL2ノルムを表す.
  評価値$E(\bm{\theta}, \bm{T})$は, $\bm{T}=\bm{T}_{target}$としたときの作業空間剛性が指令した作業空間剛性にどれだけ近いかの評価値と, 関節トルクは現状の$\bm{T}=\bm{T}_{current}$の状態から変わらないようにするための評価値を$1:\alpha$の重みで足し合わせたものである.
  集めた$N_{v1}$個のデータを$E$の値によって昇順に並べ, 上から$N_{v2}$($N_{v1}>N_{v2}$)個のデータの$\bm{T}_{rand}$の平均を$\bm{T}_{bias}$とする.
  このとき, $E(\bm{\theta}_{target}, \bm{T}_{current}+\bm{T}_{bias})$が$E(\bm{\theta}_{target}, \bm{T}_{current})$より小さい場合は$\bm{T}_{target}$を$\bm{T}_{current}+\bm{T}_{bias}$に置き換える.
  これを$N_{v3}$回繰り返し, 最終的に求まった$\bm{T}_{target}$を自己身体像に入力して所望の$\bm{l}_{target}$を得ることになる.
  ここで, $N_{v2}=1$のときは山登り法であり, 1より大きい場合はより安定に剛性の探索が可能となる.
  本研究では基本的に, $\alpha=0.02, N_{v1}=10, N_{v2}=2, N_{v3}=50$としている.
}%

\section{Experiments} \label{sec:experiment}
In all experiments, the neural networks have three layers: input, hidden, and output layers.
The number of units in the hidden layer is 1000, and the activation function is Sigmoid.

\subsection{Comparison of self-body image acquisition methods}
\begin{figure}[t]
  \centering
  \includegraphics[width=1.0\columnwidth]{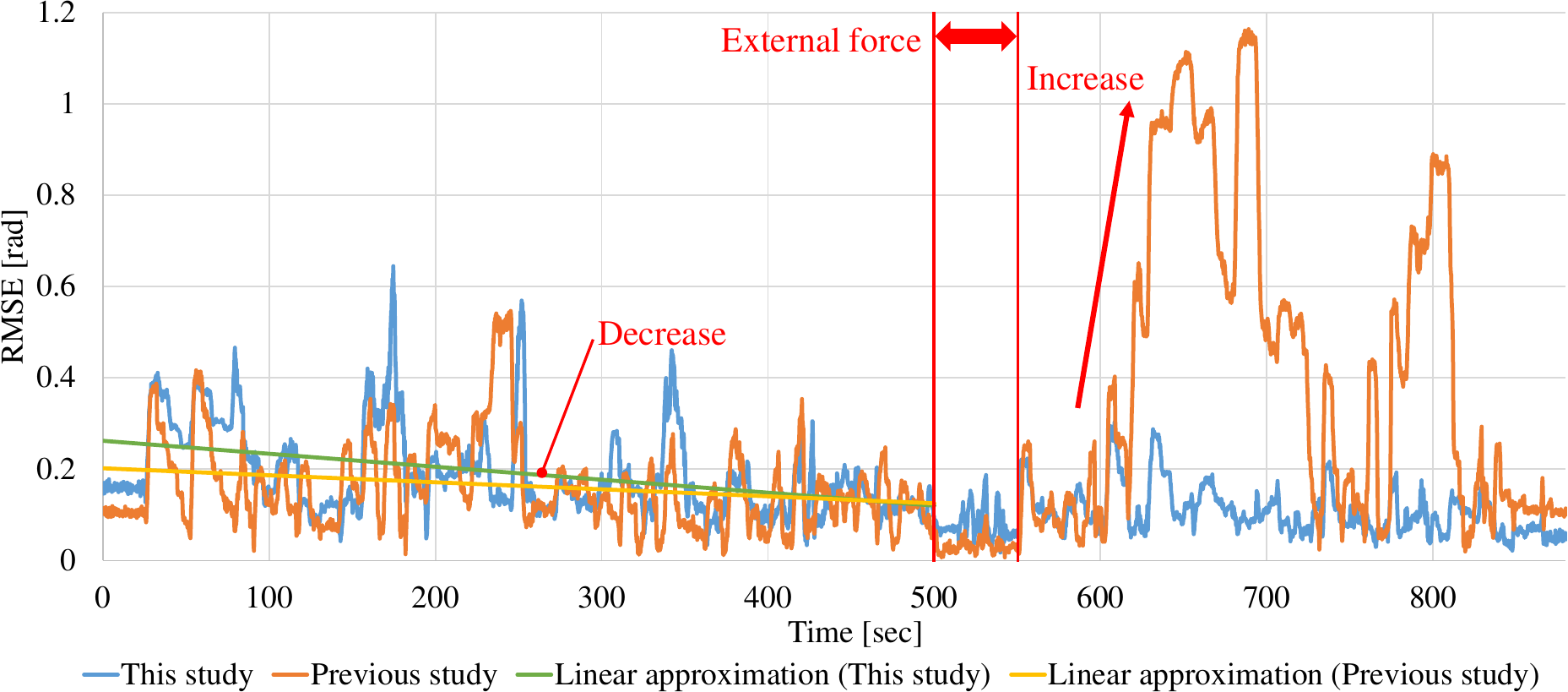}
  \caption{Comparison experiment of self-body image acquisition between the previous study and this study.}
  \label{figure:compare-experiment}
  \vspace{-3.0ex}
\end{figure}

\switchlanguage%
{%
  First, we will conduct an experiment to compare the self-body image acquisition methods of the previous study \cite{kawaharazuka2018bodyimage} and this study.
  In the previous study, self-body image was learned in a state without external force, and then was learned in a state with external force.
  In this study, first, we set the target joint angles $\bm{\theta}_{target}$ randomly in the range of the joint angle limits, set the $\bm{T}_{const}$ as $\bm{T}_{bias}$, and sent target muscle lengths obtained by \equref{eq:move-first} to the robot over 5 sec.
  Second, we sent the target muscle lengths obtained by \equref{eq:move-second} using the same target joint angles over 2 sec.
  Third, we set $\bm{T}_{const}$ as the random value from $\bm{T}_{bias}$ to $\bm{T}_{lim}$, and sent muscle lengths obtained by \equref{eq:move-first} again using the same target joint angles over 3 sec.
  By repeating these 3 steps, the space of joint angles and muscle tensions of the self-body image is learned efficiently.
  We used the 5 DOFs of the 3 DOFs shoulder and 2 DOFs elbow in Musashi for the evaluation, including a total of 10 muscles (we express muscles by the numbers $\#1\sim\#10$ shown in the right figure of \figref{figure:musculoskeletal-structure}, which include 1 polyarticular muscle).
  We applied the previous study to the right arm and applied this study to the left arm, and evaluated at the same time.

  We show the result in \figref{figure:compare-experiment}.
  RMSE is the Root Mean Squared Error of the difference between the current and estimated joint angles.
  When comparing the transition of RMSEs and their linear regressions in these two studies, both of the RMSEs decrease slowly, and the slope in this study is larger than the one in the previous study.
  This is because the two updaters of the previous study slightly compete and the self-body image is adjusted to the current sensor information too much due to lack of data accumulation.
  Also, we added external force at 500 sec.
  There is no use of torque controller.
  After that, by setting two different target joint angles, RMSE in the previous study rose rapidly.
  This is because the self-body image is adjusted to the current joint angles too much, and so the joint angle estimation goes in an unintended direction when moving to a great extent.
}%
{%
  まず, 先行研究\cite{kawaharazuka2018bodyimage}と本研究における自己身体像獲得手法の比較実験を行う.
  \cite{kawaharazuka2018bodyimage}では外力がない状態でまず学習し, その後人間が外力を加えることでオンライン学習を行っていた.
  本研究ではまず, \equref{eq:move-first}において角度限界の中でランダムに$\bm{\theta}_{target}$を指定し, $\bm{T}_{target}$は最初に一律で$\bm{T}_{bias}$を指定する.
  次に指令関節角度をそのままで\equref{eq:move-second}を実行し, 最後に, 指令関節角度はそのままで筋張力指令を$\bm{T}_{bias}$から$\bm{T}_{lim}$の間でランダムに指定し, もう一度\equref{eq:move-first}を実行する.
  この3つの手順を繰り返すことで, 効率よく自己身体像における関節角度と筋張力の空間を学習させる.

  評価には肩の3自由度と肘の2自由度の合計5自由度を用い, 筋は合計10本存在している(\figref{figure:musculoskeletal-structure}の右図における$\#1\sim\#10$の番号で表現, うち多関節が1本).
  \figref{figure:musculoskeletal-structure}のMusashiDarmの右腕に先行研究, 左腕に本研究を適用し, 同時に評価を行った.
  その結果を\figref{figure:compare-experiment}に示す.
  RMSEは現在の関節角度と推定関節角度の差の平均二乗誤差平方根(Root Mean Squared Error)を表している.

  500 secまでの両手法におけるRMSEとその線形回帰を比較すると, それぞれ徐々にRMSEが下がっていることがわかり, その傾きは本研究の方が大きいことがわかる.
  これは, 先行研究は2種類の別々の指標を学習させるupdaterにより更新されておりそれらが多少競合していること, データを蓄積・拡張していないためより現在の値に強く適合してしまうこと, 等が考えられる.
  これらにより全体の最適化がうまくできていない.

  また, 500 secのところで人間が両腕に外力を加えている.
  その後, 二回程指令関節角度を送ると, 先行研究におけるRMSEが上昇してしまっている.
  これは, ある一部の姿勢に適合しすぎたゆえ, その姿勢を大きく変更したときに関節角度推定が大きくズレてしまったからと考えられる.
}%

\subsection{Long-time Self-body Image Acquisition}
\begin{figure}[t]
  \centering
  \includegraphics[width=1.0\columnwidth]{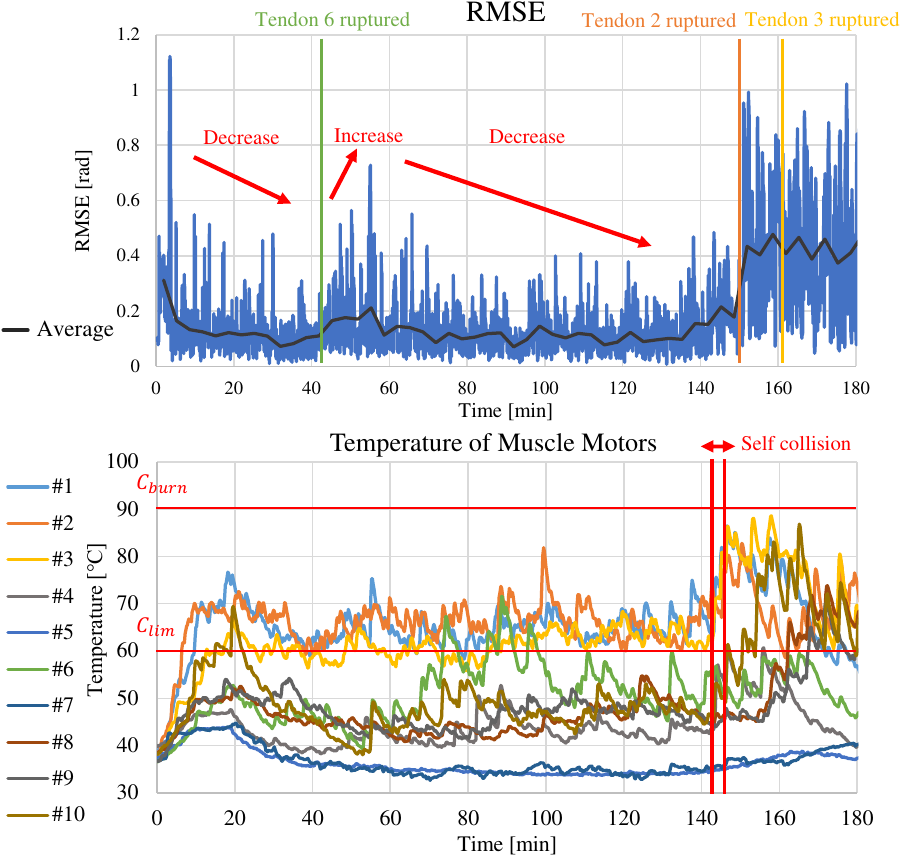}
  \caption{Transition of Root Mean Squared Error (RMSE) of the difference between target and current joint angles and muscle temperature in the long-time self-body image acquisition experiment lasting 3 hours.}
  \label{figure:longtime-experiment}
  \vspace{-3.0ex}
\end{figure}

\switchlanguage%
{%
  We conducted a long-time self-body image acquisition experiment lasting 3 hours.
  We show the result in \figref{figure:longtime-experiment}.
  In the upper figure of \figref{figure:longtime-experiment}, RMSE is the Root Mean Squared Error of the difference between the current and target joint angles, and the black line expresses the average RMSE of 4 minutes.
  We can see that the average RMSE gradually decreased from about 0.3 rad to 0.08 rad until at about 40 minutes by the online acquisition of the correct self-body image.
  Although the O-ring of muscle $\#6$ ruptured at 42 minutes and the RMSE increased to about 0.2 rad, the RMSE gradually decreased due to the redundancy of muscles and the online learning of self-body image, expressing the benefit of redundant muscles on the musculoskeletal structure well.
  After that, muscles $\#2$ and $\#3$ ruptured at about 150 minutes, the major muscles required to raise the shoulder in the roll direction vanished, and RMSE increased rapidly.
  Regarding the muscle temperatures, the temperature remained stable under 70 ${}^\circ\textrm{C}$ due to the safety mechanism of \equref{eq:safety-mechanism}.
  Also, because we did not set self-collision avoidance, at 140 minutes, target joint angles were set to values that caused the arm to sink into the self-body, and muscle temperatures rose rapidly.
  However, due to the safety mechanism, the robot could keep moving without exceeding the limit of muscle temperature $C_{burn}$ (90 ${}^\circ\textrm{C}$) which is the limit for the prevention of motor burnout.

  The main reason why the average RMSE did not fall under about 0.08 rad is the hysteresis of muscles.
  MusashiLarm has a complex muscle structure, large friction between muscles, between muscle and bone, between muscle and pulley, etc., and the final joint angles change according to the direction of movement.
}%
{%
  本節ではMusashiLarmを用いて, 3時間という長時間の間の筋骨格ヒューマノイドの自己身体像獲得実験を行う.
  \figref{figure:longtime-experiment}の上図において, RMSEは現在の関節角度と指令関節角度の差の平均二乗誤差平方根(Root Mean Squared Error)を表しており, 黒線は約4分間ごとの平均を示している.
  40分程度までは自己身体像が正しく学習されることで順調にRMSEが下がり, 平均のRMSEは0.3[rad]から0.08[rad]程度まで下がっていることがわかる.
  しかしその後, 6番の腱のOリングが切れ, RMSEは0.2[rad]程度まで上昇するが, 筋の冗長性と学習の効果により次第にRMSEは下がり, 動作は元に戻っていく.
  この筋の冗長性による動作訂正は, 筋骨格ヒューマノイドの利点を如実に見ることができる.
  その後, 150分程度で腱2と3が切れ, 肩をroll方向に上げる主要な筋群が無くなってしまったため, RMSEは急激に上昇した.
  筋の温度については, 安全機構(\equref{eq:safety-mechanism})により70度以下で安定している.
  また, 自己干渉回避等を設定していないため, 140分の時に自身の体にめり込むような指令が発生し急激な温度上昇が起こっているが, 安全機構によって, 焼損を防ぐために設定している温度限界$C_{burn}$(90度)を越えずに動作し続けることができている.

  RMSEが0.08[rad]程度までしか下がらない最も大きな理由はヒステリシスであると考えている.
  複雑な構成ゆえに筋同士や筋と骨格, 筋とプーリ等における摩擦が大きく, 動く方向によって最終的な関節角度が変わってしまう.
}%

\subsection{Dumbbell Raise Experiment}
\begin{figure}[t]
  \centering
  \includegraphics[width=0.9\columnwidth]{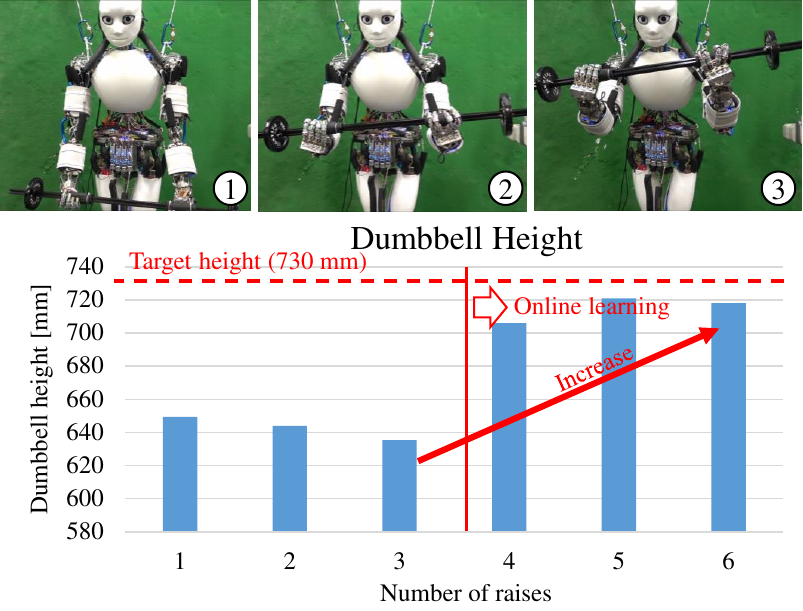}
  \caption{Result of dumbbell raise experiment. Lower graph expresses the transition of the dumbbell height.}
  \label{figure:dumbbell-experiment}
  \vspace{-3.0ex}
\end{figure}

\switchlanguage%
{%
  We conducted a dumbbell (3 kg) raise experiment by position control using the self-body image, while running its online learning.
  We show the appearance in the upper figure of \figref{figure:dumbbell-experiment}, and show the transition of the dumbbell height in the lower figure of \figref{figure:dumbbell-experiment}.
  When raising the dumbbell several times, the relationship of muscle lengths, muscle tension, etc. of the dumbbell raise is incorporated into the self-body image, so Musashi was gradually able to raise the dumbbell to the target height of 730 mm.
}%
{%
  自己身体像を用いた位置制御により, オンライン学習を実行しながらダンベルを上げる実験を行う.
  \figref{figure:dumbbell-experiment}の上図にその様子を, 下図にダンベルの高さ変化を示す.
  ダンベルを何度か上げると, オンライン学習により, その動作の際に必要な筋張力, 筋長等の関係が分かっていき, 徐々にターゲットである730 mmに近づいていることがわかる.
}%

\subsection{Variable Stiffness Control Using Self-body Image} \label{subsec:vsc-exp}
\begin{figure}[t]
  \centering
  \includegraphics[width=1.0\columnwidth]{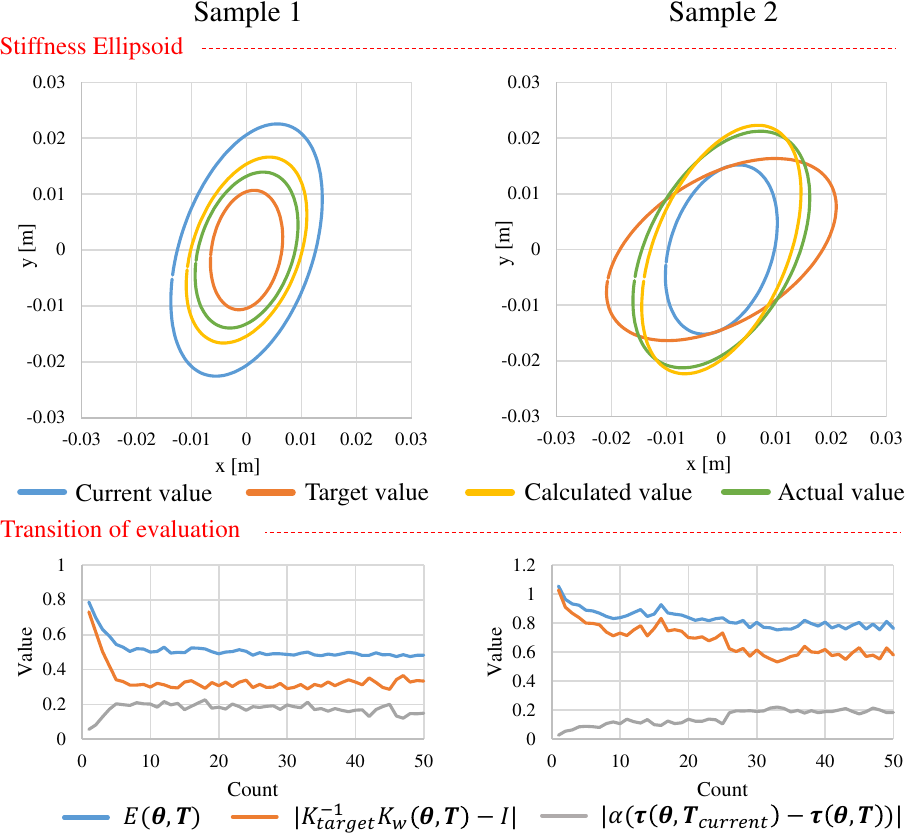}
  \caption{Stiffness ellipsoid and transition of evaluation in the evaluation experiment of variable stiffness control using self-body image.}
  \label{figure:variable-stiffness-control-experiment}
  \vspace{-3.0ex}
\end{figure}

\switchlanguage%
{%
  First, we conducted the evaluation of the variable stiffness control.
  We set $\bm{\theta}_{target}$ as $({S}_{r}, {S}_{p}, {S}_{y}, {E}_{p}, {E}_{y})=(45, 0, 0, -90, 0)$ [deg], and conducted experiments to change the operational stiffness as intended ($S$ means the shoulder, $E$ means the elbow, and $rpy$ means the roll, pitch, and yaw rotation. These symbols are also used in subsequent experiments).
  The result is shown in \figref{figure:variable-stiffness-control-experiment}.
  Each stiffness ellipsoid in the upper graphs of \figref{figure:variable-stiffness-control-experiment} represents the operational displacement of the hand when $|\bm{F}|=10$ [N].
  In Sample 1, we set the target stiffness $K_{target}$ (Target Value) as a stiffness twice that of the current stiffness (Current Value), and searched for target muscle tensions by the method of \secref{subsec:variable-stiffness}.
  The transition of $E(\bm{\theta}, \bm{T})$ in Sample 1 when searching for target muscle tensions is shown in the lower graph of \figref{figure:variable-stiffness-control-experiment}, showing that this method could make the self-body stiffness close to the target stiffness, while inhibiting the error between current and calculated joint torques (in this graph, we display the value $E$ if $E(\bm{\theta}_{target}, \bm{T}_{current}+\bm{T}_{bias})>E(\bm{\theta}_{target}, \bm{T}_{current})$).
  When we move the actual robot using $\bm{T}_{target}$, which realizes the final stiffness (Calculated Value) calculated by the method, we can observe the final current stiffness of the actual robot (Actual Value).
  In Sample 1, $\bm{T}_{target}$, which realizes the target stiffness, could not be obtained, but this method succeeded in making the current stiffness close to the target stiffness.
  Also, the stiffness ellipsoids of Calculated Value and Actual Value were almost the same, and it shows that the self-body image was learned correctly.
  In Sample 2, we set the target stiffness by changing the scale and slope of the current stiffness ellipsoid.
  As a result of the search, the same scale of the target stiffness ellipsoid was realized, but the slope was not, because there was no solution to realize the target stiffness and the method selected the best choice.
}%
{%
  まず, 身体剛性の可変制御に関する評価を行う.
  $({S}_{r}, {S}_{p}, {S}_{y}, {E}_{p}, {E}_{y})=(45, 0, 0, -90, 0)$ [deg]を$\bm{\theta}_{target}$に設定し, 剛性を変化させる実験を行った($S$は肩, $E$は肘, $rpy$はroll, pitch, yawを表す).
  この結果を\figref{figure:variable-stiffness-control-experiment}に示す.
  まずSample 1では, 現状の身体剛性(Current value)から長軸短軸に関して剛性が2倍になるような$K_{target}$(Target value)を設定し, \secref{subsec:variable-stiffness}によって現在の身体剛性を指令剛性に近づけるような探索を行った.
  探索の際の$E(\bm{\theta}, \bm{T})$の推移はSample 1の下図のようになっており, 関節トルク誤差を抑えつつ, 身体剛性を近づけることができているのがわかる.
  その結果得られた身体剛性(Calculated value)を実現する$\bm{T}_{target}$と自己身体像を用いて実機を動作させた結果, 最終的な現状態での身体剛性(Actual value)が観測される.
  Sample 1では, 本研究の手法によって, 指令した剛性楕円体に完全に一致した剛性楕円体を実現する$\bm{T}_{target}$は探索できなかったものの, 指令値に大きく近づけることに成功している(剛性楕円体は, $|\bm{F}|=10$ [N]のときの変位として記述している).
  また, その$\bm{T}_{target}$を指令として実機を動作させた後における剛性楕円体は, 探索された剛性楕円体とほぼ一致しており, 自己身体像の学習が正しくできていることがわかる.
  Sample 2においては, 大きさと傾きを変えた剛性楕円体を指令値としたが, 探索の結果, 大きさは同等なものが実現できたが, 傾きには誤差が残ってしまった.
  これは, どんなに探索してもその指令剛性を実現できる解がなく, その中でも最良の選択をした結果だと考える.
}%

\begin{figure}[t]
  \centering
  \includegraphics[width=1.0\columnwidth]{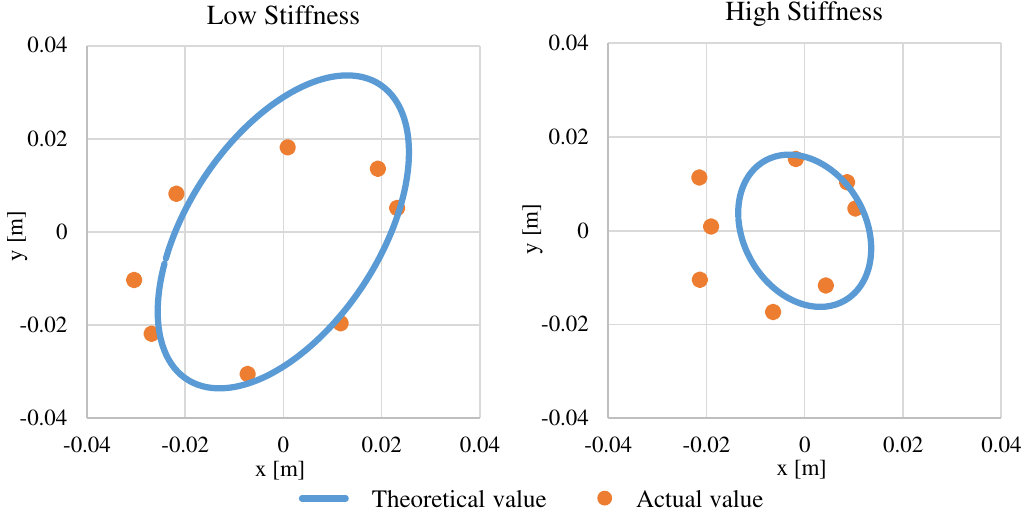}
  \caption{Stiffness ellipsoid in the evaluation experiment of variable stiffness estimation using self-body image.}
  \label{figure:variable-stiffness-evaluation-experiment}
  \vspace{-3.0ex}
\end{figure}

\switchlanguage%
{%
  Second, we verified to what extent the theoretical stiffness ellipsoid calculated by the method in \secref{subsec:variable-stiffness} matches the actual stiffness ellipsoid.
  We set $\bm{\theta}_{target}$ as $({S}_{r}, {S}_{p}, {S}_{y}, {E}_{p}, {E}_{y})=(0, 30, 0, -60, 0)$ [deg], which is the posture in which the body stiffness can be measured easily, and created a state of low stiffness and high stiffness by the method in \secref{subsec:variable-stiffness}.
  Then, we added 10 N force to the end effector on the $xy$ plane, while measuring the value by a forcegauge, from 8 directions equally dividing 360 deg, and measured the displacement of the end effector by potentiometers.
  The result is shown in \figref{figure:variable-stiffness-evaluation-experiment}.
  When the body stiffness is low or high, we can see that the theoretical and actual ellipsoid match to a certain degree.
  The error between the theoretical and actual value is considered to be due to the remaining difference between the acquired self-body image and the actual robot, and the hysteresis by friction as above.
}%
{%
  次に, \secref{subsec:variable-stiffness}によって計算される作業空間剛性の理論値が現実とどの程度合致するかについて検証する.
  身体剛性を測定しやすいよう, $({S}_{r}, {S}_{p}, {S}_{y}, {E}_{p}, {E}_{y})=(0, 30, 0, -60, 0)$ [deg]を$\bm{\theta}_{target}$に設定し, 剛性が低い状態, 剛性が高い状態を\secref{subsec:variable-stiffness}の手法により作り出す.
  このとき, xy平面においてエンドエフェクタに対して, 周回360度を均等に分割した8方向から, フォースゲージの測定値が10 Nになるように力を加え, その時のエンドエフェクタのxy平面における変位を計測した.
  この結果を\figref{figure:variable-stiffness-evaluation-experiment}に示す.
  剛性が高いときも低いときも, 理論値と実際の値がある程度合致していることがわかった.
  この結果における理論値と実際の値の誤差は, 学習された自己身体像と実機の間に残った差異と, 上記にも述べた摩擦等によるヒステリシスが問題となっていると考えられる.
}%

\subsection{Impact Correspondence Experiment}
\begin{figure}[t]
  \centering
  \includegraphics[width=0.95\columnwidth]{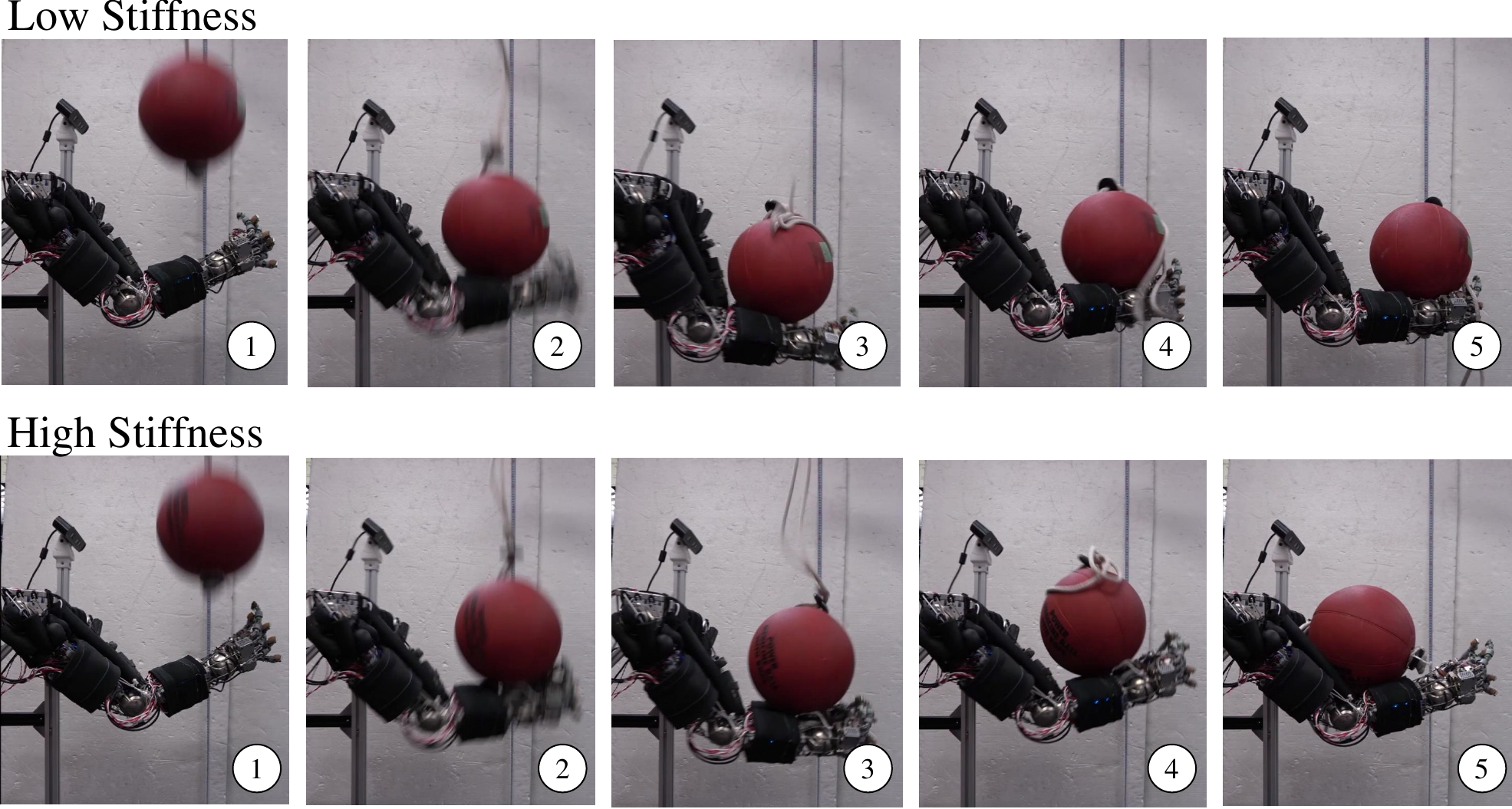}
  \caption{Impact correspondence experiment with variable stiffness control.}
  \label{figure:catch-experiment}
  \vspace{-3.0ex}
\end{figure}

\begin{figure}[t]
  \centering
  \includegraphics[width=1.0\columnwidth]{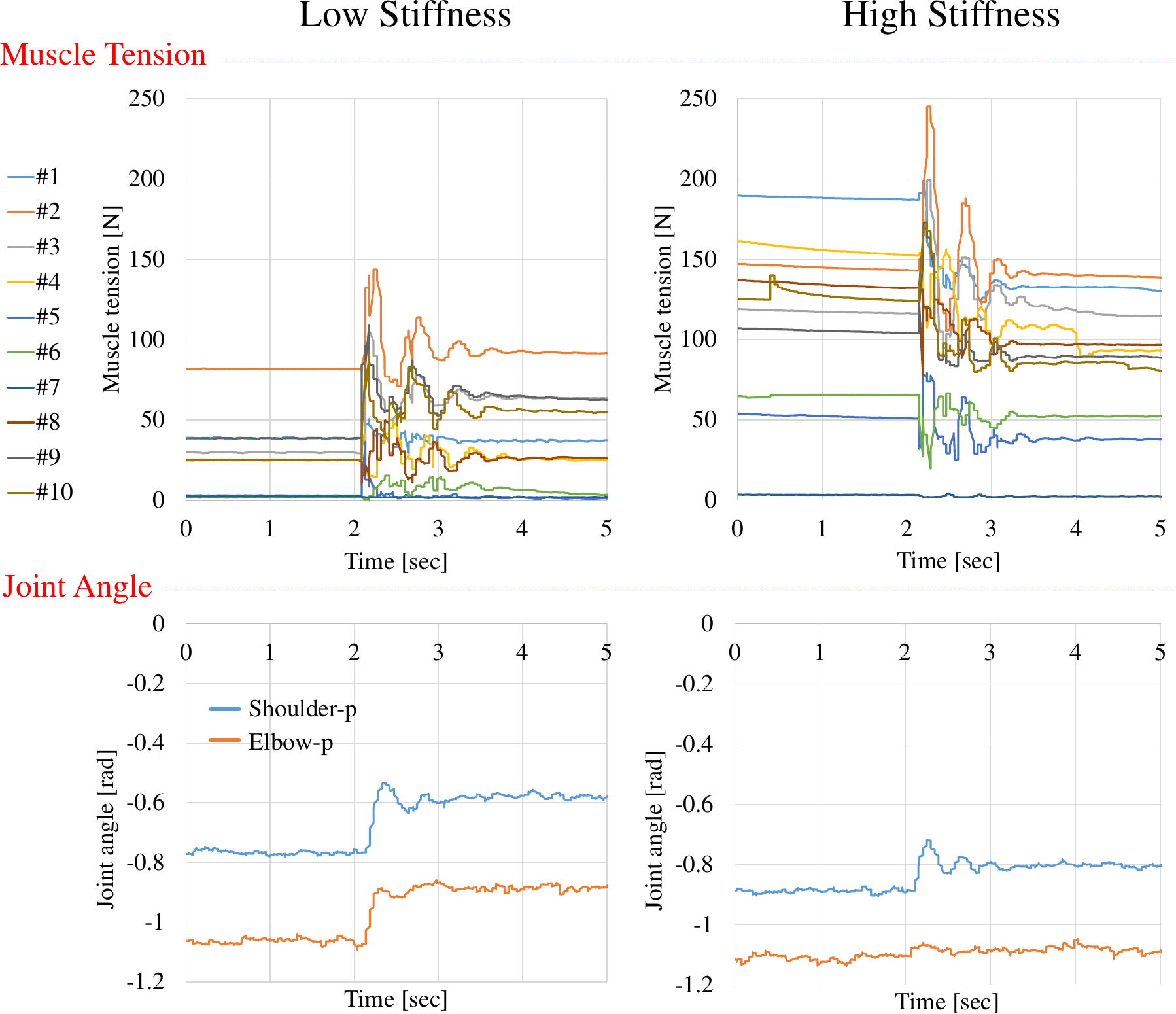}
  \caption{Transition of muscle tensions and joint angles in the impact correspondence experiment with variable stiffness control.}
  \label{figure:catch-experiment-graph}
  \vspace{-3.0ex}
\end{figure}

\switchlanguage%
{%
  We conducted an impact correspondence experiment using variable stiffness control.
  The joint angles were set as $({S}_{r}, {S}_{p}, {S}_{y}, {E}_{p}, {E}_{y})=(-45, -30, -20, -60, 0)$ [deg], a 5 kg ball was dropped from 1 m above and 0.15 m in front of the elbow with the hands clasped together, and the transitions of muscle tensions and joint angles between low stiffness and high stiffness were compared.
  The movement of the dual arm in this experiment is shown in \figref{figure:catch-experiment}, and the transitions of muscle tensions and joint angles of the left arm are shown in \figref{figure:catch-experiment-graph}.
  From \figref{figure:catch-experiment}, we can see that the arms moved to a great extent when the stiffness was low, and the displacement of the arms was not as large when the stiffness was high.
  Additionally, in the lower graph of \figref{figure:catch-experiment-graph}, we can see that both displacements of joint angles in Shoulder-p and Elbow-p become small in the high stiffness state.
  Also, in the upper figure of \figref{figure:catch-experiment-graph}, although the evaluation of muscle tensions is difficult because the initial muscle tensions are different between low and high stiffness, the maximum muscle tension is about 150 N in low stiffness and is about 250 N in high stiffness, indicating that the low stiffness state can absorb sudden impact.
}%
{%
  可変剛性制御を用いた衝撃対応実験を行う.
  ここでは, \secref{sec:musculoskeletal-structure}で説明した筋骨格上肢\cite{kawaharazuka2018musashilarm-en}の左腕を対称にしてい右腕を作成し, 双腕にしたものを用いる.
  この関節角度を$({S}_{r}, {S}_{p}, {S}_{y}, {E}_{p}, {E}_{y})=(-45 -30 -20 -60 0)$ [deg]に設定し, 手同士を握った状態において, 肘上1 m, 肘前0.15 mから50 Nのボールを落とし, そのときの筋張力・関節角度の遷移を, 低剛性・高剛性において比較実験する.
  実験の際の筋骨格双腕の動きを\figref{figure:catch-experiment}に, 左手における筋張力・関節角度の遷移を\figref{figure:catch-experiment-graph}に示す.
  \figref{figure:catch-experiment}において, 明らかに低剛性のときは大きく手が動き, 高剛性のときは関節角度の変位が低剛性のときほど大きくないことがわかる.
  実際に, \figref{figure:catch-experiment-graph}の下図において, Shoulder-p, Elbow-p両方において高剛性にすることで衝撃に対する変位を小さくすることができていることがわかる.
  また, その際の筋張力変化(\figref{figure:catch-experiment-graph}の上図)においては, 初期の筋張力が低剛性と高剛性で大きく違うため評価は難しいものの, 低剛性では最大150 N程度, 高剛性では最大250 N程度と, 低剛性の方が衝撃を吸収できていることもわかる.
}%

\section{CONCLUSION} \label{sec:conclusion}
\switchlanguage%
{%
  In this study, we proposed a method for long-time self-body image acquisition, and position, torque, and variable stiffness control using the self-body image.
  For long-time self-body image acquisition, we considered simplifying the online updater for stable learning, accumulating and augmenting the actual robot sensor information without wasting it, and including a safety mechanism to inhibit high muscle tension and temperature, and we succeeded in conducting a 3 hour learning experiment.
  Also, by using the self-body image, we realized position control using 2 stage feedback of muscle tension, torque control using muscle Jacobian calculated from the differentiation of the self-body image, and variable stiffness control using hill-climb method.

  In future works, we would like to propose further structures of self-body image considering the hysteresis of muscles and dynamic movements.
}%
{%
  本研究では筋骨格ヒューマノイドにおける長期的な逐次的自己身体像獲得手法の提案, また, その自己身体像を用いた位置・トルク・可変剛性制御について提案した.
  長期的に自己身体像を獲得するためには, 学習を安定させるためにシンプルな更新則にすること, データを無駄にせず蓄積・拡張して用いること, 筋温度や筋張力の高まりを抑制する安全機構が必要であると考え, 実装を行い, 3時間に及ぶ自己身体像獲得実験を行った.
  また, 自己身体像を用いて, 二段階の筋張力フィードバックにより位置制御を実現できること, 自己身体像を微分して筋長ヤコビアンを得ることでトルク制御を実現できること, 山登り法により可変剛性制御が実現できることを確認した.

  今後は, 本研究の実験から必要と感じた, 自己身体像における動的特性や筋のヒステリシスを考慮したより正しい自己身体像の構成・学習手法の開発を進めたい.
}%

{
  \bibliographystyle{IEEEtran}
  \bibliography{main}
}

\end{document}